\documentclass{article}
\usepackage[normalem]{ulem}
\usepackage{amsmath}
\usepackage{color}
\usepackage{array}
\usepackage{stfloats}
\usepackage{amsthm} 
\usepackage{multirow}
\usepackage{amssymb}
\usepackage{float}
\usepackage{nccmath}
\usepackage{graphicx}
\usepackage{setspace}
\usepackage{booktabs}
\usepackage{mwe}
\usepackage{cite}
\usepackage{amsmath,amssymb,amsfonts}
\usepackage{graphicx}
\usepackage{textcomp}
\usepackage{xcolor}
\usepackage{physics}
\usepackage{bm}
\usepackage[utf8]{inputenc}

\usepackage[algo2e]{algorithm2e} 
\usepackage{xspace}
\usepackage{subfigure}
\usepackage{hyperref}
\usepackage{kotex}

\newcommand{\BfPara}[1]{{\noindent\bf#1.}\xspace}
\usepackage[accepted]{icml2021}

\icmltitlerunning{Communication and Energy Efficient Slimmable Federated Learning via Superposition Coding and Successive Decoding}

\begin{document}

\twocolumn[
\icmltitle{Communication and Energy Efficient Slimmable Federated Learning via Superposition Coding and Successive Decoding}

\icmlsetsymbol{equal}{}

\begin{icmlauthorlist}
\icmlauthor{Hankyul Baek}{kor,equal}
\icmlauthor{Won Joon Yun}{kor,equal}
\icmlauthor{Soyi Jung}{kor}
\icmlauthor{Jihong Park}{deakin}
\icmlauthor{Mingyue Ji}{utah}
\icmlauthor{Joongheon Kim}{kor}
\icmlauthor{Mehdi Bennis}{oulu}
\end{icmlauthorlist}%

\icmlaffiliation{kor}{Korea University, Seoul 02841, Korea.}
\icmlaffiliation{deakin}{Deakin University, Geelong, VIC 3220, Australia.}
\icmlaffiliation{utah}{The University of Utah, UT 84112, USA.}
\icmlaffiliation{oulu}{University of Oulu, Oulu 90014, Finland}%

\icmlcorrespondingauthor{Jihong Park}{jihong.park@deakin.edu.au}
\icmlcorrespondingauthor{Joongheon Kim}{joongheon@korea.ac.kr}%

\icmlkeywords{Machine Learning, ICML}

\vskip 0.3in
]

\printAffiliationsAndNotice{\icmlEqualContribution{:Hankyul Baek and Won Joon Yun.\\}} 

\begin{abstract}
Mobile {devices} are indispensable sources of big data. Federated learning (FL) has a great potential in exploiting these private data by exchanging locally trained models instead of their raw data. However, mobile {devices} are often energy limited and wirelessly connected, and FL cannot cope flexibly with their heterogeneous and time-varying energy capacity and communication throughput, limiting the adoption. Motivated by these issues, we propose a novel energy and communication efficient FL framework, coined \emph{SlimFL}. To resolve the heterogeneous energy capacity problem, each {device} in SlimFL runs a width-adjustable \emph{slimmable neural network (SNN)}. To address the heterogeneous communication throughput problem, each full-width ($1.0$x) SNN model and its half-width ($0.5$x) model are \emph{superposition-coded} before transmission, and \emph{successively decoded} after reception as the $0.5$x or $1.0$x model depending on the channel quality. Simulation results show that SlimFL can simultaneously train both $0.5$x and $1.0$x models with reasonable accuracy and convergence speed, compared to its vanilla FL counterpart separately training the two models using $2$x more communication resources. Surprisingly, SlimFL achieves even higher accuracy with lower energy footprints than vanilla FL for poor channels and non-IID data distributions, under which vanilla FL converges slowly.
\end{abstract}\section{Introduction}\label{sec:1}

Federated learning (FL) is a promising solution to enable high-quality on-device learning at mobile devices such as phones \cite{FLGoogle18}, cars \cite{Sumudu:GC18}, and drones \cite{Hamid:TCOM20}. Each of these devices has only a limited amount of local data, and FL can overcome the lack of local model training samples by exchanging and aggregating the local models of different devices. To reach its full potential, it is essential to scale up the range of federating devices that are often wirelessly connected while having heterogeneous levels of available energy~\cite{Google:FL19,pieee21park}. This mandates addressing the following interrelated energy and wireless communication problems.

\begin{figure}[t!]
\centering
    \subfigure[Uplink.]{\includegraphics[width=.495\columnwidth]{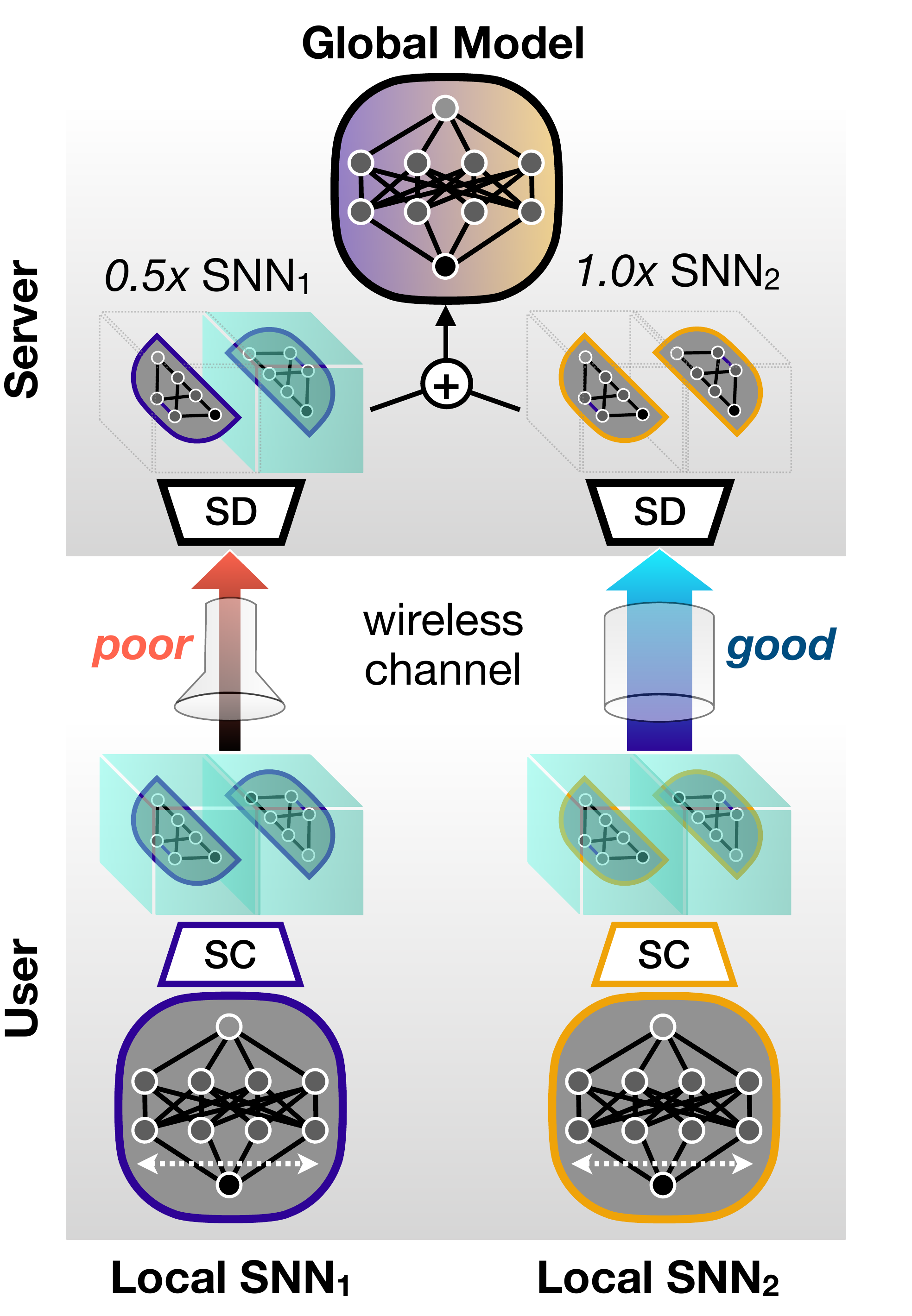}}
    \subfigure[Downlink.]{\includegraphics[width=.495\columnwidth]{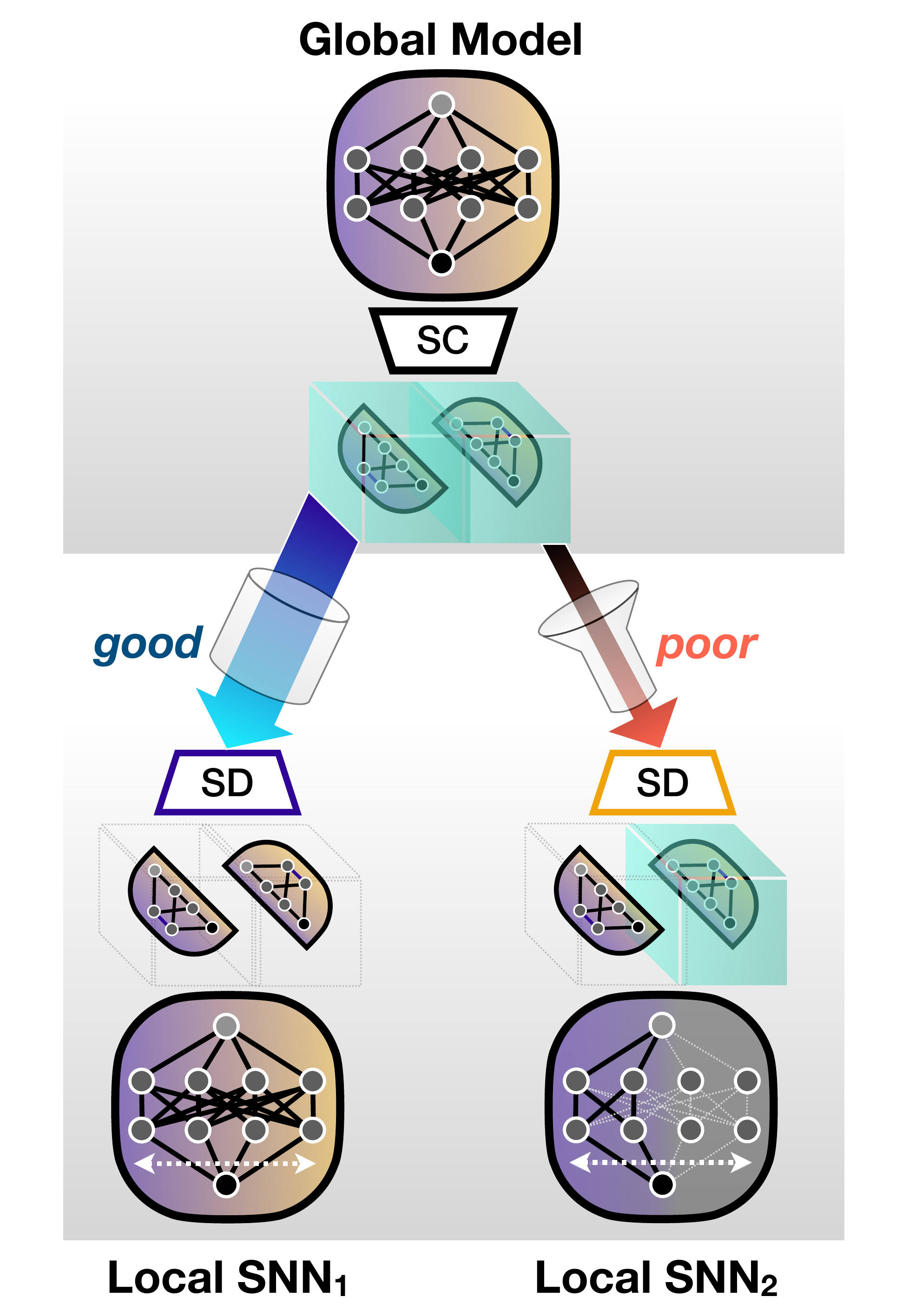}}
    \caption{Slimmable federated learning (SlimFL) in the (a) uplink and (b) downlink, with the superposition coding (SC) and successive decoding (SD) of the slimmable neural networks (SNNs).}
    \label{fig:abstract}
    \vspace{-5mm}
\end{figure}

\BfPara{Heterogeneous Energy Capacity Problem} Different devices have heterogeneous levels of available energy. Low-energy devices are likely to run small models, whereas high-energy devices prefer to operate large models. Unfortunately, FL can only aggregate the local models under the same architecture \cite{Brendan17}, so is able to train either small or large models at a time. To cope with heterogeneous energy capacity, one should therefore perform FL two times with the increased overall training time, or run FL simultaneously for two separate groups of devices with reduced training samples while compromising accuracy. Performing FL using a width-controllable \emph{slimmable neural network (SNN)} architecture enables to train these two-level models at once while federating across all devices, after which each trained local SNN model can adjust its width due to its desired energy consumption \cite{Arxiv2018_Slimmable,ICCV2019_USlimmable}.

\BfPara{Heterogeneous Communication {Throughput Problem}}
Unlike wired connectivity, wireless communication channel conditions vary across different devices. When the channel information is known before transmission, poor-channel devices can only exchange small models, while good-channel devices can participate in FL using large models. SNNs allows them to collaborate together, in a way that poor-channel devices send their local SNNs after reducing the widths, and contribute only to a fraction of the entire global model construction. This however entails extra communication and energy costs for probing channel conditions \cite{4267831} that change over time and locations due to random fading and device mobility \cite{TseBook:FundamaentalsWC:2005}. 

Spurred by the aforementioned problems, in this article we propose the first \emph{SNN-based FL algorithm} that leverages \emph{superposition coding (SC)} and \emph{successive decoding (SD)}, coined \emph{slimmable FL (SlimFL)}. By applying SNNs to FL, SlimFL can address the heterogeneous energy {capacity} problem. Besides, by exploiting SC and SD, SlimFL can proactively cope with the heterogeneous channel {throughput} problem for unknown channel state information. 

To illustrate, consider an SNN with two width levels as shown in Fig.~\ref{fig:abstract}. In the uplink from each device to the server, the device uploads its local updates after jointly encoding the left-half (LH) and the right-half (RH) of its local SNN model while allocating different transmission power levels to them, i.e., SC \cite{Cover:TIT72}. Then, the server first attempts to decode the LH. If decoding the LH is successful, the server successively tries to decode the RH, i.e., SD or also known as successive interference cancellation (SIC) \cite{Haenggi:TWC12}. Consequently, when the device-server channel {throughput} is low, the server can decode only the LH of the uploaded model, obtaining the \emph{half-width (0.5x)} model. When the channel {throughput} is high, the server can decode both LH and RH, and combine them to yield the \emph{full-width model (1.0x)}. The same principle is also applied in the downlink from the server to each device.

For the two-level SNN architecture under MNIST and Fasion-MNIST classification tasks, simulations results corroborate that SlimFL can simultaneously train both 0.5x and 1.0x models each of which achieves reasonable accuracy and training speed, compared to the vanilla FL baselines having fixed 0.5x and 1.0x widths that should be separately trained without applying SC and SD while incurring $1.5$x higher communication costs. What is more, under poor communication channels and non-independent and identically distributed (non-IID) data distributions, SlimFL achieves higher accuracy and lower energy costs (thanks to faster convergence) than its vanilla FL counterpart whose training becomes unstable in the same conditions.
\section{Related Work}\label{sec:2}

\BfPara{Depth/Width-Adjustable Neural Networks} 
To meet different on-device energy and memory requirements, it is common to prune model weights \cite{Han:16,seo2021communicationefficient} or transfer a large trained model's knowledge into a small empty model via knowledge distillation (KD) \cite{HintonKD:14,seo2020federated}, which however incurs additional training operations. Alternatively, one can adjust a trained model's width and/or depth in accordance with the resource requirements. Following this principle, depth-controlled neural networks \cite{IJCNN2019_DepthControllable} and adaptive neural networks \cite{AAAI2019_Anytime} can adjust their depths after training, whereas SNNs tune their widths \cite{Arxiv2018_Slimmable,ICCV2019_USlimmable}. In this paper, we leverage width-controllable SNNs, and develop its FL version, SlimFL. Such an extension is non-trivial, and entails several design issues, such as local SNN training algorithms, aggregating segment prioritization (e.g., more aggregating the same LH/RH segments vs. balancing the LH and RH aggregations), which will be discussed in Sec~\ref{sec:4}.

\BfPara{Superposition Coding and Successive Decoding}
Since its invention by Thomas Cover in 1972 \cite{Cover:TIT72}, SC has been widely utilized in communication systems, particularly for simultaneously supporting different devices in the context of non-orthogonal multiple access (NOMA) \cite{7842433}. In a nutshell, SC encodes two different data signals into one while allocating two different power levels before transmissions. After receptions, SD decodes the SC-encoded signal by first decoding the stronger signal, followed by subtracting it and decoding the remainder as the weaker signal \cite{TseBook:FundamaentalsWC:2005}. The same principle is also applicable for supporting a single device simultaneously requesting two types of data with different priorities \cite{ParkGC:18}, such that the higher priority signal should almost surely be decoded while the lower priority signal can be successively decoded only under good channel conditions. Inspired by this, SlimFL makes an SNN's LH a higher priority so as to receive the $0.5$x model even under poor channels. It can decode the SNN's RH only when the channel conditions are good, obtaining the $1.0$x model by combining both LH and RH. Consequently, SlimFL ensures stable convergence under poor channels, which will be studied in Sec.~\ref{sec:5}.

\section{Preliminaries: SC and SD}\label{sec:3}
\BfPara{Decoding Success Probability}
The successful reception of a wireless signal is mainly affected by the signal-to-interference-plus-noise ratio (SINR) \cite{TseBook:FundamaentalsWC:2005}. At a receiver, the SINR is given as
$\gamma={g d^{-\beta} P^T }/{(\sigma^2 + P^I)}$,
where $P^T$, $P^I$, and $\sigma^2$ stand for the transmission power, the received interference power, and the noise power. The $d$ is the transmitter-receiver distance, $\beta\geq 2$ is the path loss exponent, and $g$ is a random small-scale fading gain. Following the Shannon's capacity formula with a Gaussian codebook~\cite{Shannon}, the received throughput $R$ with the bandwidth $W$ is given as:
\begin{align}
    R = W \log_2(1 + \gamma) \quad \text{(bits/sec)} .
  \label{eq:bitrate}
\end{align}
When the transmitter encodes raw data with a code rate~$t$, its receiver can successfully decode the encoded data if $R \geq t$. The decoding success probability is derived~as:
 \begin{align}
    \Pr(R \geq t)&= \Pr(\frac{g d^{-\beta} P^T }{\sigma^2 + P^I} \geq t'), \label{eq:dsp1}
    \end{align}
where $t'=2^{\frac{t}{W}}-1$.

\BfPara{Superposition Coding (SC)} 
We consider simultaneously conveying $K$ messages from a transmitter to its receiver. These messages are SC-encoded before transmission \cite{Cover:TIT72}, while the total transmission power budget $P^T$ is allocated to the $k$-th message with the amount of $P^T = \sum\limits^{K}_{k=1}\nolimits P^T_{k}$ transmission power for $k\in[1,K]$.
When conveying only a single message, i.e., $K=1$, there exists no interference at reception, i.e., $P^I=0$. For $K>1$, SD determines the interference as elaborated next.

\BfPara{Successive Decoding (SD)} At its receiver, the SC-encoded $K$ messages are supposed to be successively reconstructed by first decoding the stronger signal, followed by cancelling out the reconstructed (stronger) signal and then decoding the next stronger signal, i.e., SD, also known as successive interference cancellation \cite{Haenggi:TWC12,Jinho:TCOM17}. Assuming $P_k^T > P_{k'}^T$ for $k'>k$, the receiver can successively decode the $k$-th message while experiencing the rest of the messages as its interference $P_k^I$, i.e.,

\begin{align}
    P^I_k = g d^{-\beta} \hat{P}_k^I,  \label{eq:interference}
\end{align}
where $\hat{P}_k^I:=\sum^{K}_{k' = k+1}P^T_{k'}$ for $k\leq K-1$, and $\hat{P}^I_K=P^I_K=0$ as there is no interference for the last message. 
Let $R_k$ denote the throughput for the $k$-th message. By substituting  \eqref{eq:interference} into \eqref{eq:dsp1}, the distribution of $R_k$ is cast as, $\Pr(R_k \geq t ) = \Pr\left(g \geq \frac{c}{ P^T_k/t' - \hat{P}^I_k}      \right)$,
where $c=\sigma^2 d^\beta$.
Applying this result, the decoding success probability $p_k$ of the $k$-th message is derived as:
\begin{align}
    &p_k 
    = \Pr(R_1 \geq t, R_{2} \geq t,\cdots, R_k \geq t  )\nonumber\\
    &\hspace{-10pt} = \Pr\left( g \!\geq\!  \max \!\left\{\!
    \frac{c}{ P^T_1/t' - \hat{P}^I_1} , \cdots, \frac{c}{ P^T_k/t' - \hat{P}^I_k} 
    \!\right\}\!\right). \label{eq:sicdsp11}
\end{align}
The derivation details of \eqref{eq:sicdsp11} are deferred to Appendix~\ref{sec:appendix_dsp}.

In SlimFL, as Fig.~\ref{fig:abstract} shows, we consider $K=2$ while treating the LH and RH segments of each model as $2$ messages. SC and SD are applied in both uplink (from each device to the server) and downlink (from the server to each device) to be elaborated in Sec.~\ref{sec:4}. The impact of SC and SD on each uplink or downlink is studied in Appendix~\ref{sec:appendix_E}. 

\section{SlimFL with SC and SD}\label{sec:4}
In this section, we elaborate the architectures and training algorithms of SlimFL. The network under study consists of $N$ devices. Following the federated averaging algorithm~\cite{Brendan17}, the $n$-th device trains a local SNN model $\theta^n$ using its local dataset, and communicates with a parameter server storing a global model. The SNN $\theta^n$ is divided into the LH segment $\theta^n_{1/2}$ and the RH segment $\theta^n_{2/2}$, such that the $0.5$x model is $\theta^n_{1/2}$ and the $1.0$x model is the stack of both segments, i.e., $\theta^n = \textsf{Stack}\{\theta^n_{1/2}, \theta_{2/2}\}$. 

In the uplink, the device uploads the SC-encoded local model $\theta^n$ to the server. After reception, according to \eqref{eq:sicdsp11} with $K=2$ and $P_1^T\gg P_2^T$, the server can successively decode using SD and obtain: (i) the $1.0$x model if the channel fading gain satisfies {\small$g\geq  \max\{ c/(P_1^T/t' - P_2^T), c/(P_2^T/t') \}$\normalsize}; (ii) the $0.5$x model if {\small$g\geq  c/(P_1^T/t' - P_2^T)$\normalsize}; and (iii) otherwise it obtains no model. In the downlink, SC and SD are also applied, so the global model is obtained at each device according to the aforementioned conditions.

In between the uplink and downlink, the device locally trains its SNN. Existing SNN architectures and training algorithms are intended for standalone learning~\cite{ICCV2019_USlimmable}. Towards improving the scalability for federated learning, we propose a lighter model architecture and training algorithm as detailed next.

\begin{table}[t!]
    \small \centering
    \resizebox{\columnwidth}{!}{\begin{tabular}{l|r}
    \toprule[1pt]
      \bf{Description}                & \bf{Value}  \\ \midrule
        Learning rate ($\eta$)        & $10^{-3}$ \\
        Optimizer                     & Adam\\ 
        Distance  ($d$)               &   1\\
        Path loss exponent ($\beta$)  & 2.5 \\
        Bandwidth per device ($W$)& $1.15\times10^5$ [Hz] \\
        Uplink transmission power  ($P^T_{up}$) & 25 [mW] \\
        Downlink transmission power ($P^T_{dn}$) & 100 [mW]\\
        Noise power ($\sigma_g^2$) -- good case &  $-80.6~\mathrm{[dB/Hz]}$\\
        Noise power ($\sigma_p^2$) -- poor case & $-90.6~\mathrm{[dB/Hz]}$\\\midrule[1pt] 
      \bf{UL--MobileNet Layers}             & \bf{Weight connection of layers 1.0x (0.5x)}  \\ \midrule
      \multicolumn{1}{l|}{\bf{Convolution layer}} &$C_{in}\times C_{out} \times K \times K$\\
      Conv2D + ReLU6 & $1 \times 32 \times 3 \times 3~(1 \times 16 \times 3 \times 3)$  \\
      Conv2D + ReLU6 & $1 \times 32 \times 3 \times 3~(1 \times 16 \times 3 \times 3)$  \\
      Conv2D + ReLU6 & $32 \times 32 \times 1 \times 1~(16 \times 16 \times 1 \times 1)$\\
      Conv2D + ReLU6 & $1 \times 32 \times 3 \times 3~(1 \times 16 \times 3 \times 3)$  \\
      Conv2D + ReLU6 & $32\times 64 \times 1 \times 1~(16 \times 32 \times 1 \times 1)$ \\
      \multicolumn{1}{l|}{\bf{Fully connected layer}}& $C_{in}\times C_{out}$ \\
      Linear         & $64 \times 10~(32 \times 10)$ \\ \bottomrule[1pt] 
    \end{tabular}}
    \caption{Parameters and model architecture of UL--MobileNet, where $C_{in}$, $C_{out}$, $K$, and ReLU6 stand for dimension of input channel, dimension of output channel, kernel size, and $\text{ReLU}6(x) = \min(\max(0, x), 6)$, respectively.}
    \label{tab:tab_parameters}
    \vspace{-4mm}
\end{table}

\BfPara{Ultra Light MobileNet} 
The state-of-the-art SNN architecture is the {US-MobileNet} proposed in~\cite{ICCV2019_USlimmable}. As opposed to a \emph{de facto} standard neural network architecture with a universal batch normalization (BN) layer, US-MobileNet is equipped with multiple separate BN layers to cope with all slimmable model configurations. While effective in standalone learning, in SlimFL with wireless connectivity, not all slimmable model segments are exchanged due to insufficient communication throughput, while the exchanged model segements are aggregated across devices, diluting the effectiveness of BN. In our experiments we even observed training convergence failures due to BN. Furthermore, managing multiple BN layers not only consumes additional memory costs, but also entails high computing computing overhead. For these reasons, we remove BN layers, and consider a lighter version of US-MobileNet, named \emph{Ultra Light MobileNet (UL-MobileNet)}, with the specifics provided in Tab.~\ref{tab:tab_parameters}. Compared to US-MobileNet with more than 100M FLOPS, UL-MobileNet costs only {2.76M FLOPS.}

\begin{figure}[t!]
    \centering
    \includegraphics[width=\columnwidth]{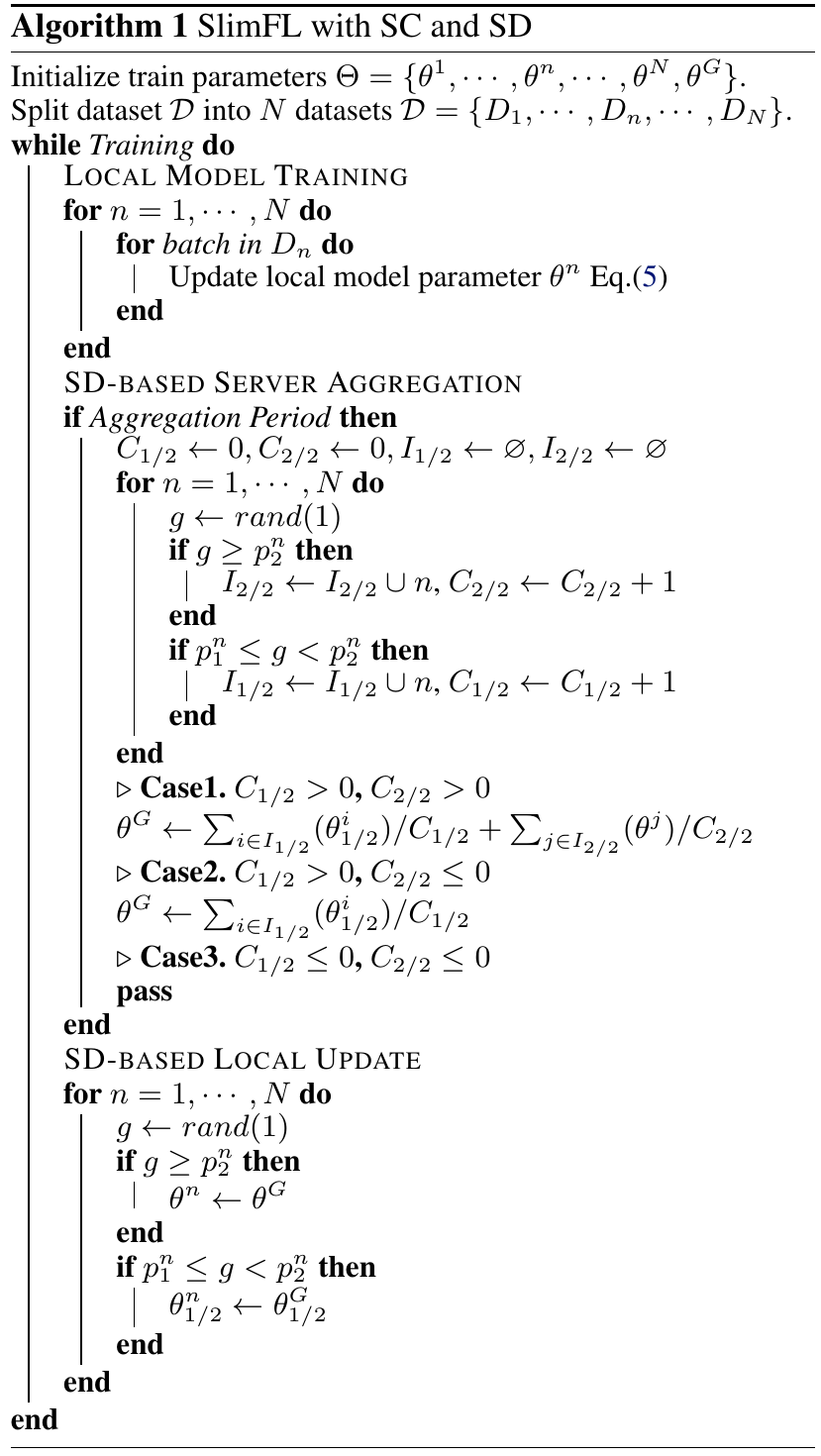}
\end{figure}

\BfPara{Superposition Training} 
The sandwich rule and the inplace distillation algorithm are two notable techniques for training an SNN~\cite{ICCV2019_USlimmable}, where the $1.0$x model becomes a teacher guiding its sub-width models via knowledge distillation. By nature, it gives more benefits under a larger SNN (i.e., a better teacher) having more sub-width configurations (i.e., more students). This is unfit for our case with the UL-MobileNet having only two width configurations, the $0.5$x and $1.0$x models. Furthermore, distillation requires to sequentially train the teacher and student models, incurring additional memory and computing costs. Alternatively, inspired a technique for a depth-controllable neural network \cite{AAAI2019_Anytime}, we consider \emph{superposition training} in which the cross-entropy loss $l(\theta^n_{1/2})$ for the $0.5$x model and the loss $l(\theta^n)$ for the $1.0$x model are weighted averaged, yielding the following weight update rule:
\begin{align}
    \theta^n \leftarrow  \theta^n - \eta \left(w_1 \nabla_{\theta^n_{1/2}}l(\theta^n_{1/2}) + w_2 \nabla_{\theta^n} l(\theta^n) \right),
 \label{eq:localtrain}
\end{align}
where $w_1$ and $w_2$ are positive constants, and $\eta$ is a learning rate. With such lighter training overhead, in our SlimFL experiments, superposition training achieved even slightly higher accuracy than that with the sandwich rule and inplace distillation, as further discussed in Appendix~\ref{sec:Appendix-1}. The overall SlimFL operations are summarized in Algorithm~1.
\section{Evaluation}\label{sec:5}
To show the effectiveness and feasibility of SlimFL, in this section we present the performance of SLimFL exploiting SC and SD compared to its Vanilla FL counterpart without SC nor SD, in terms of accuracy, communication efficiency, and energy efficiency, as well as their robustness to various channel conditions and non-IID data distributions.

\BfPara{Baselines}
Our goal is enabling each device to obtain both large and small models so as to cope with its large and small energy levels in future. To this end, by leveraging SNNs with SC and SD, SlimFL simultaneously exchanges and trains $0.5$x and $1.0$x models by consuming the per-device bandwidth $W$, uplink transmission power $P_{up}^T$, and downlink transmission power $P_{dn}^T$. This is compared with a Vanilla FL baseline, \emph{Vanilla FL-$1.5$x}. Due to the lack of width-adjustable SNNs, each device in Vanilla FL-$1.5$x separately runs fixed-width $0.5$x and $1.0$x models, referred to as \emph{Vanilla FL-$0.5$x} and \emph{Vanilla FL-$1.0$x}, respectively. Without SC nor SD, the device exchanges both $0.5$x and $1.0$x models separately. In brief, Vanilla FL-$1.5$x is tantamount to simultaneously running the two federated averaging operations separately for $0.5$x and $1.0$x models by doubling the bandwidth, transmission power, and computing resources. For clarity, we report the performance of Vanilla FL-$0.5$x and Vanilla FL-$1.0$x individually if available (i.e., accuracy, received bits), and otherwise we report only Vanilla FL-$1.5$x (i.e., energy cost).

\begin{table}[t!]
    \small \centering
    \resizebox{\columnwidth}{!}{\begin{tabular}{c|l|cc}
    \toprule[1pt]
      \multicolumn{2}{c|}{\bf{Description}}  & \bf{1.0x} & \bf{0.5x}  \\ \midrule
     \multirow{3}{*}{Computation}  &  MFLOPS / round & $2.76$ & $0.79$ \\ 
       &  \# of parameters & $4,586$ & $2,293$\\
       &  Bits / round & $172,688$ & $86,344$ \\\midrule
        Transmission & Uplink w. SC+SD & $5.0$ [mW] & $20.0$ [mW] \\ 
       Power ($P^T$)& Downlink, w. SC+SD & $20.0$ [mW] & $80.0$ [mW] \\
       \bottomrule[1pt]    
    \end{tabular}}
    \caption{Computing costs of UL-MobileNet and transmission power in the uplink and downlink.}
    \label{tab:tab_cost}
    
    \vspace{-4mm}
\end{table}

\begin{figure*}

\small\centering
\begin{tabular}{ccc}
    \includegraphics[width=0.28\linewidth]{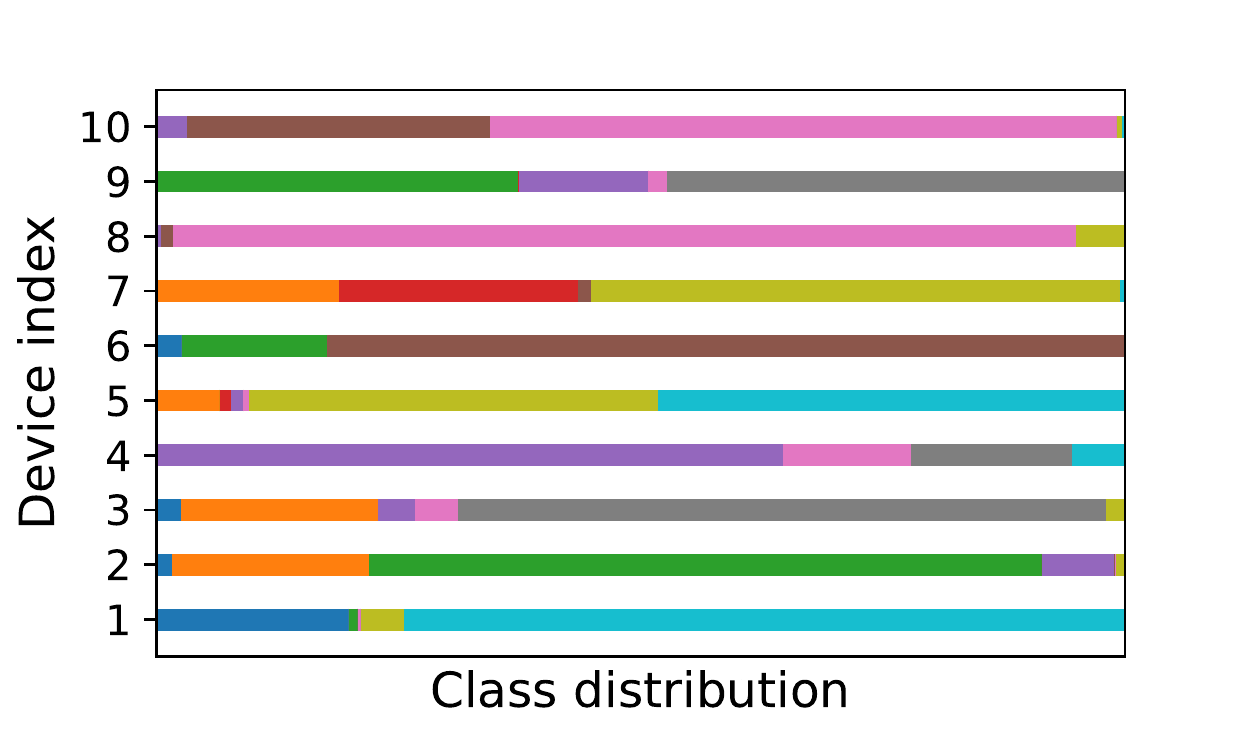} & \includegraphics[width=0.28\linewidth]{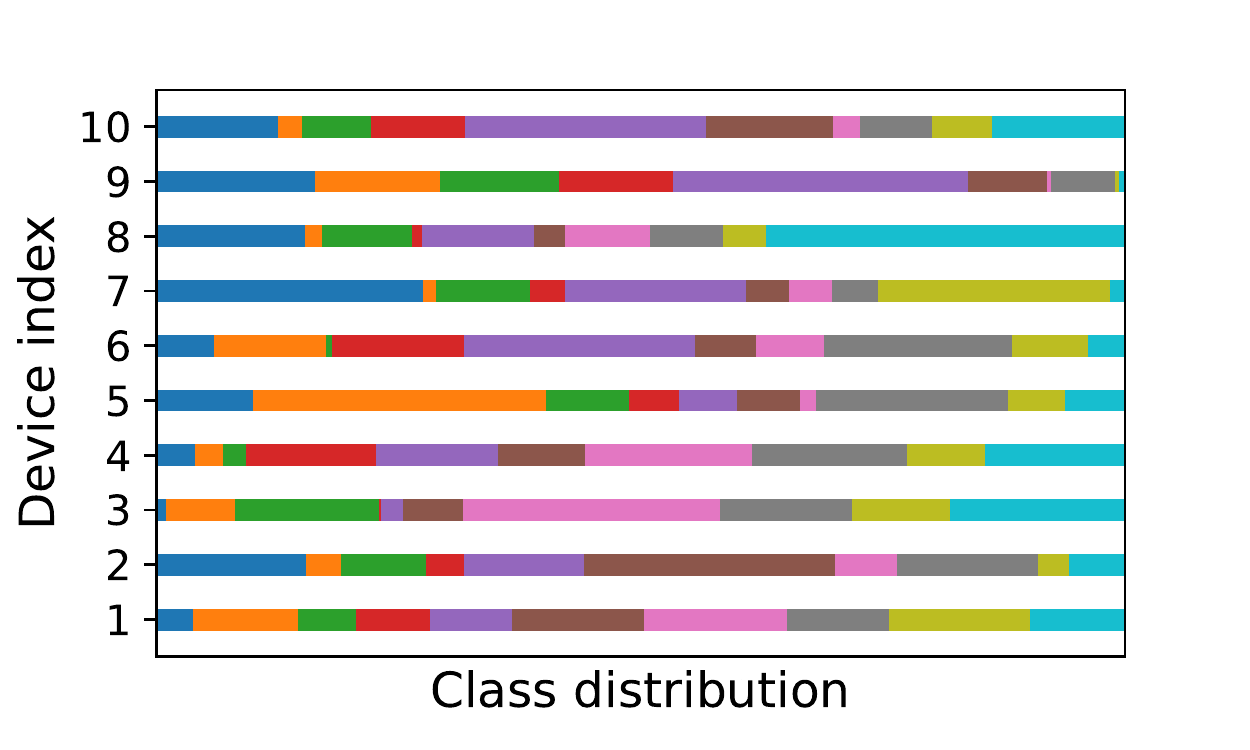} & 
    \includegraphics[width=0.28\linewidth]{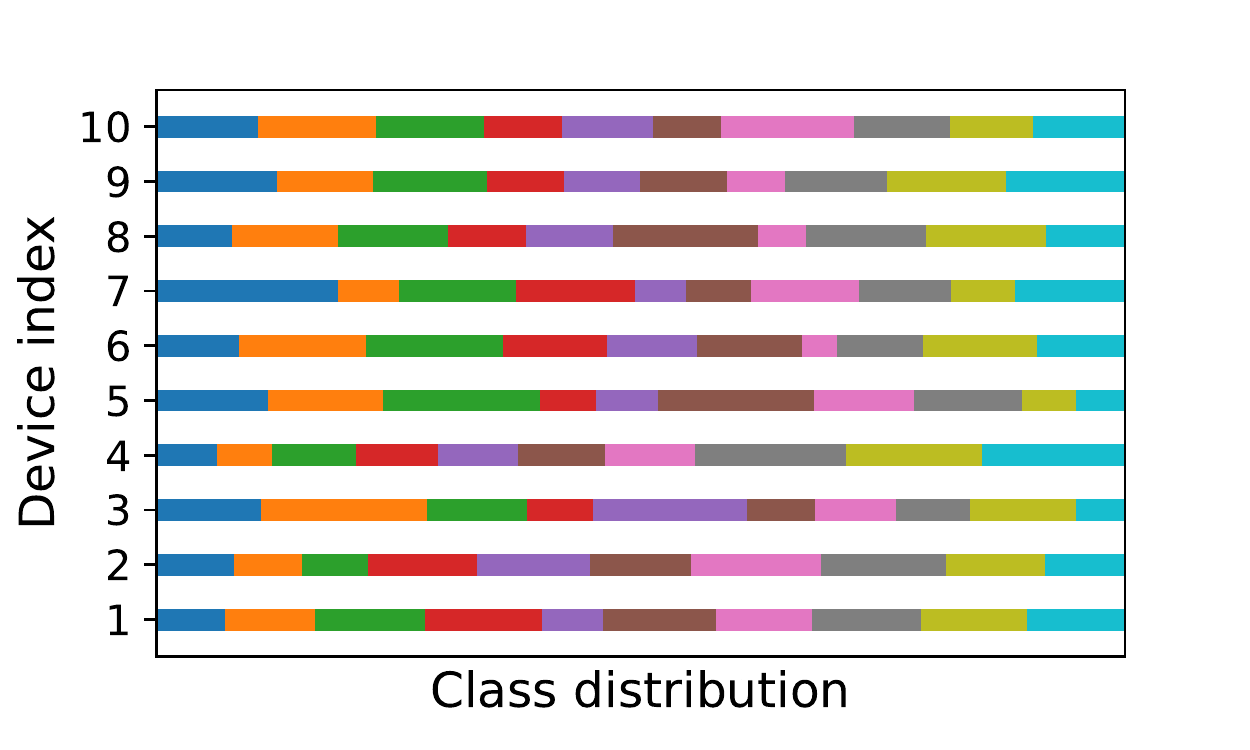}\\
    (a) $\alpha = 0.1$ (extremely non-IID). & (b) $\alpha = 1.0$. & (c) $\alpha = 10$ ($\approx$ IID). \\ 
\end{tabular}
    \caption{Data distributions across devices for the different values of the Dirichlet concentration ratio $\alpha$.}
    \label{fig:Non-iid-distribution}
    \vspace{-4mm}
\end{figure*}

\BfPara{Simulation Settings}

We consider a classification task by default with the Fashion MNIST dataset. Following the method proposed in~\cite{Arxiv_2019_NonIID}, the non-IIDness of the dataset distribution across devices is controlled by the Dirichlet distribution with its concentration parameter $\alpha\in\{0.1, 1.0, 10\}$, where a lower $\alpha$ is more non-IID distributed (i.e., more imbalanced numbers of samples over labels across devices), as visualized in Fig.~\ref{fig:Non-iid-distribution}. For a limited set of simulations, the MNIST dataset is also studied, which is deferred to Appendix~\ref{sec:appendix_dataset}. A single round of uplink and downlink communications is followed by every single local training epoch. {The communication channels over different devices are orthogonal in both uplink and downlink.} The small-scale fading gain $g$ for each channel realization follows an exponential distribution $g\sim\textsf{Exp}(1)$, i.e., Rayleigh fading \cite{TseBook:FundamaentalsWC:2005}. Other communication and training hyperparameters as well as the SNN model architecture are summarized in Tab.~\ref{tab:tab_parameters}. For the given transmission power levels and channel environments, the decoding success probabilities in the uplink and downlink are provided in Tab.~\ref{tab:my_label}, which is based on \eqref{eq:sicdsp11} and calculated using \eqref{eq:p1exp} and \eqref{eq:p2exp} in Appendix~\ref{sec:appendix_dsp}.

\begin{table}[t!]
    \small\centering
    \resizebox{\columnwidth}{!}{\begin{tabular}{c|cccccccc}
        \toprule[1pt]
        \multirow{1}{*}{\bf{Channel}} & \multicolumn{8}{c}{\bf Decoding Success Probability}   \\
         \bf{Condition} & $p^{up}_1$ & $p^{up}_2$ &  $p^{up}_{cp1}$ & $p^{up}_{cp2}$ & $p^{dn}_{1}$ & $p^{dn}_{2}$& $p^{dn}_{cp1}$ & $p^{dn}_{cp2}$ \\ \midrule
        Good & 0.983 & 0.964 & 0.993 & 0.973 & 0.996 & 0.991 & 0.998 & 0.993  \\ 
        Poor & 0.810 & 0.632 & 0.912 & 0.704 & 0.948 & 0.891 & 0.977 & 0.916  
        \\ \bottomrule[1pt]
    \end{tabular}}
    
    \caption{Decoding success probability under different channel conditions.}
    \vspace{-3mm}
    \label{tab:my_label}
\end{table}

\begin{figure*}[t!]
\small\centering
\begin{tabular}{ccc}
    \includegraphics[width=0.291\linewidth]{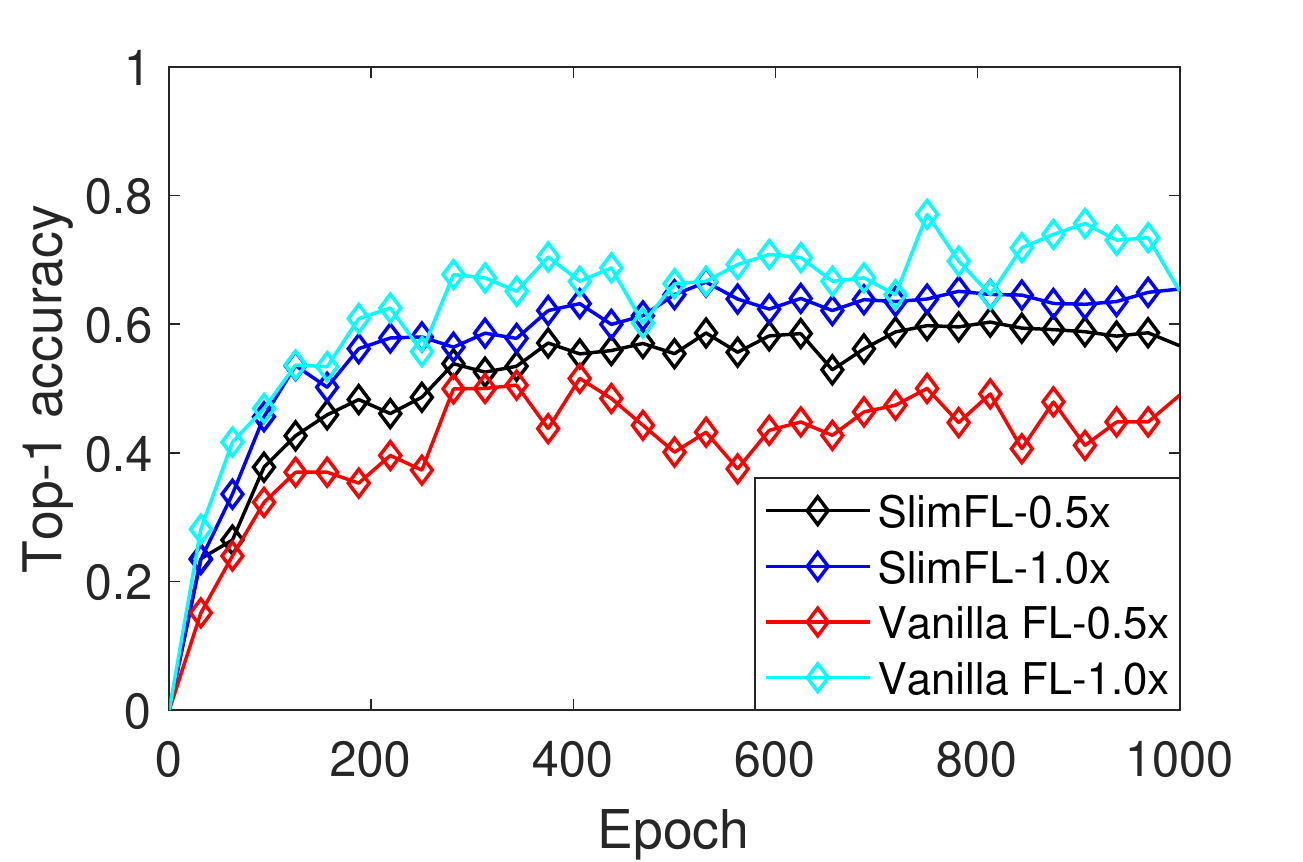} &
    
    \includegraphics[width=0.291\linewidth]{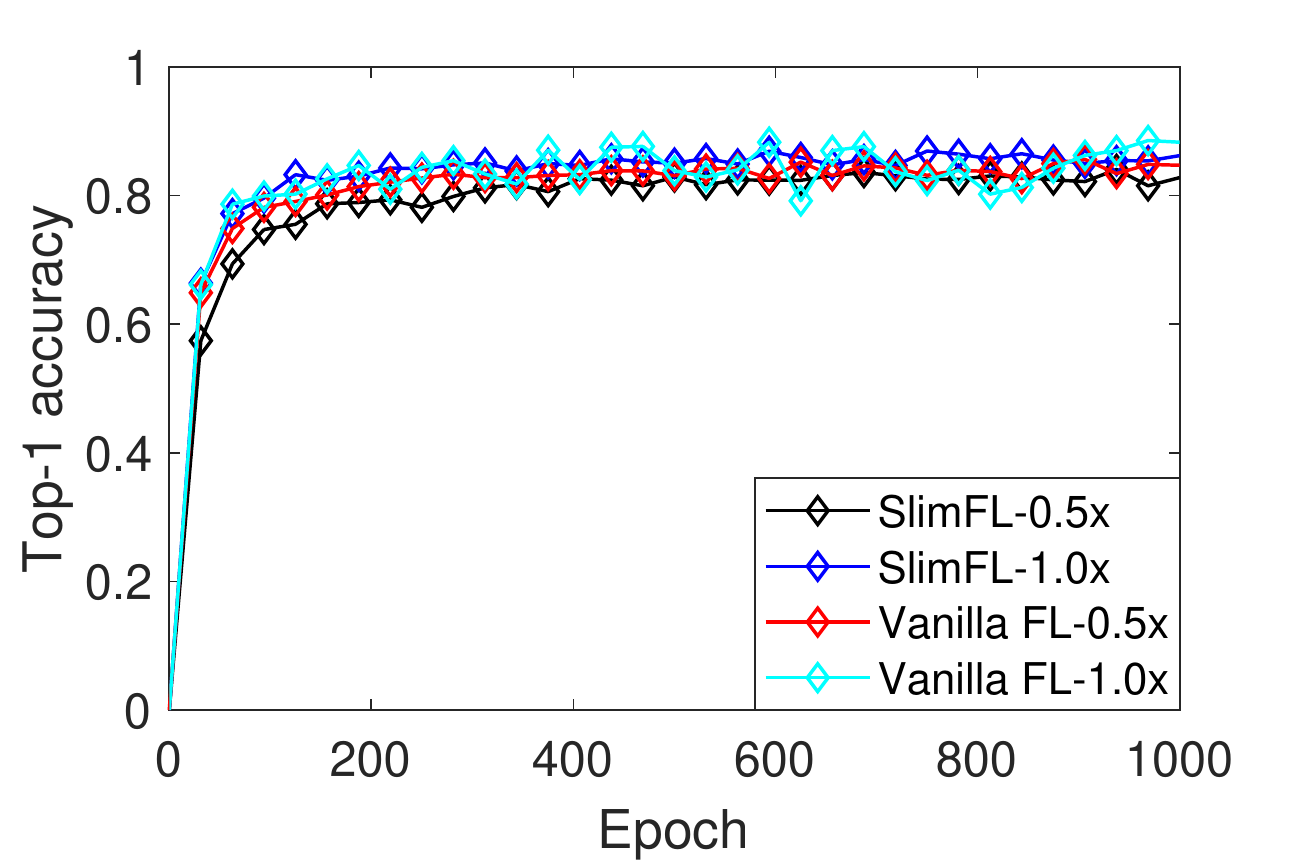} & 
    
    \includegraphics[width=0.291\linewidth]{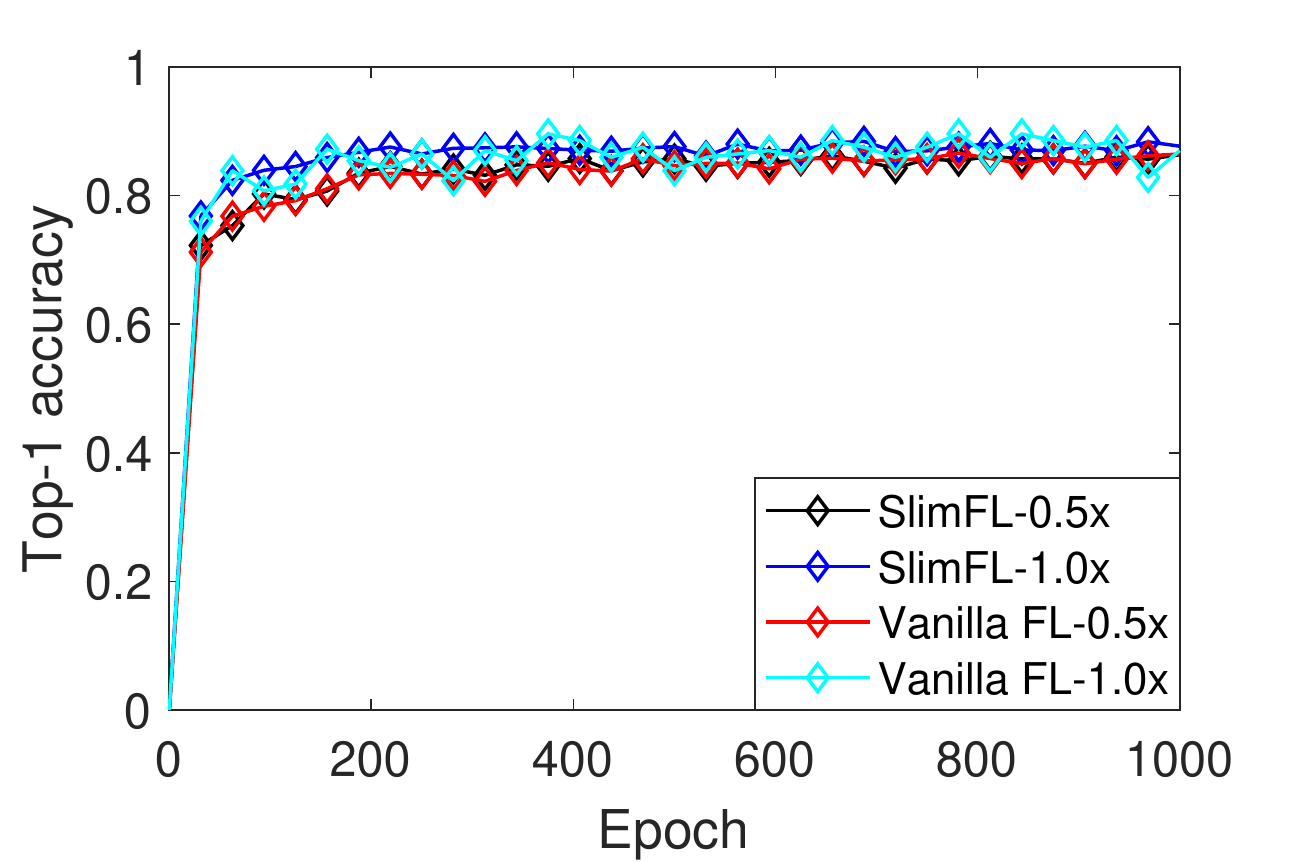}\\
    (a) $\alpha = 0.1$ & (b) $\alpha = 1.0$ & (c) $\alpha = 10$ \\ 
\end{tabular}
    \vspace{-2.5mm}
    \caption{{Test accuracy} in good channel conditions (on average).}
    \label{fig:performance-good}
\end{figure*}
\begin{figure*}[t!]
\small\centering
\begin{tabular}{ccc}
       \includegraphics[width=0.291\linewidth]{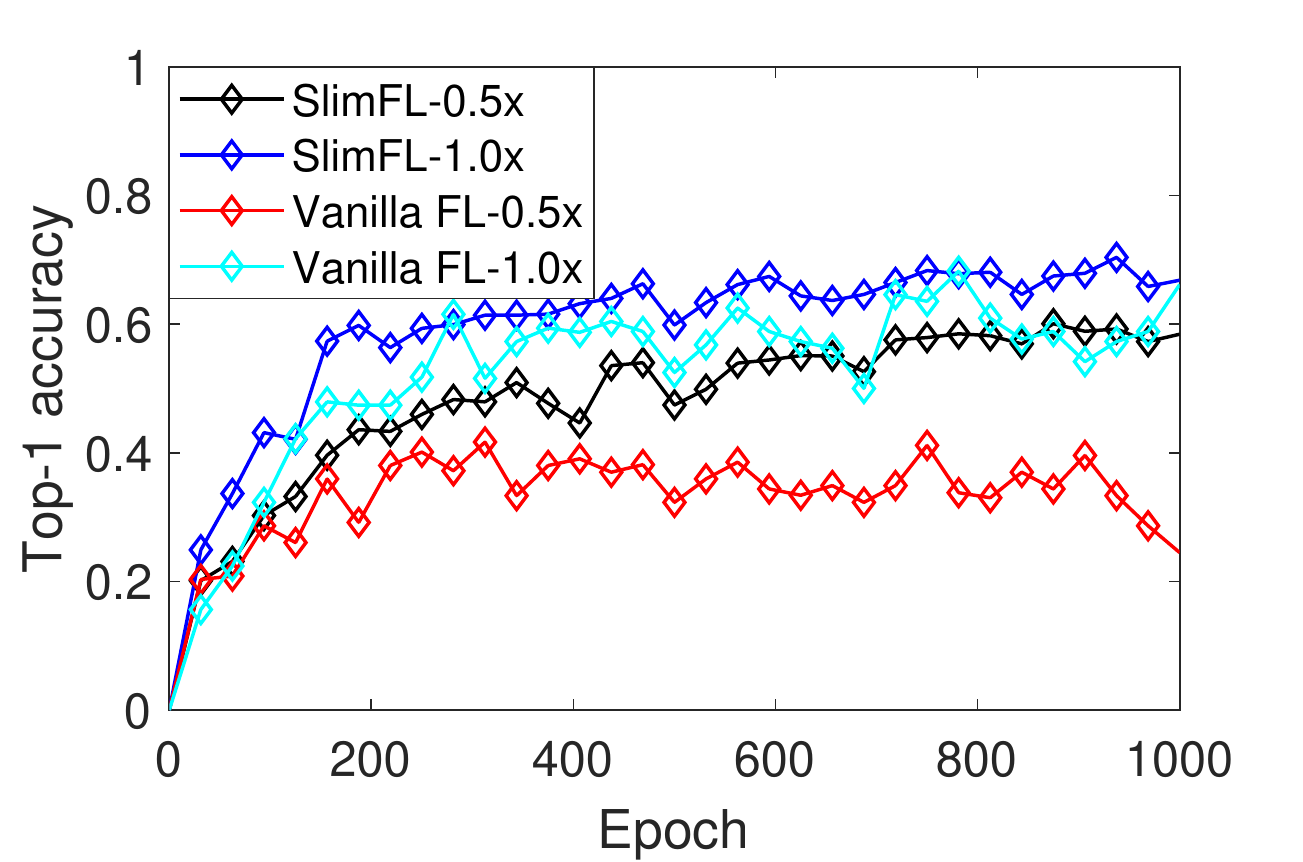} &
       \includegraphics[width=0.291\linewidth]{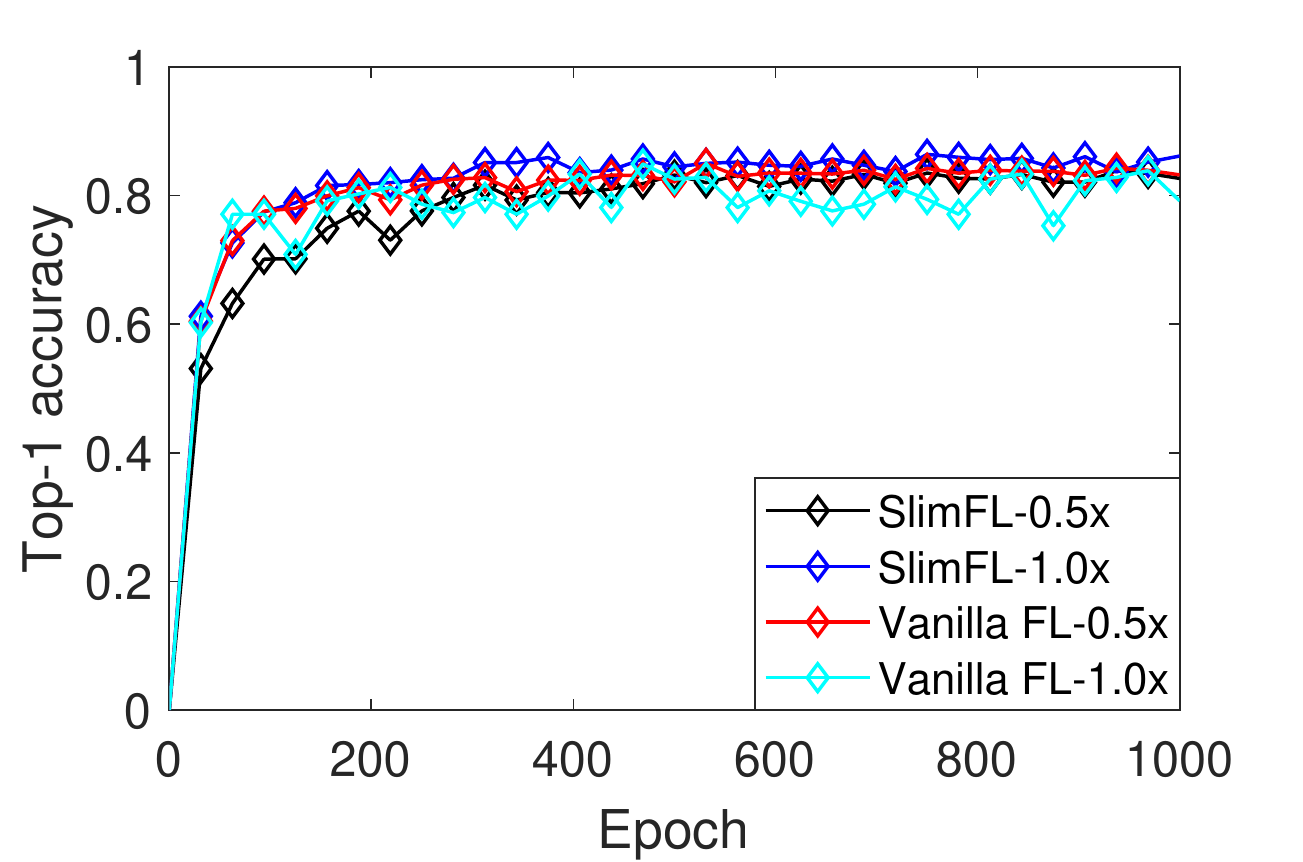} & 
    \includegraphics[width=0.291\linewidth]{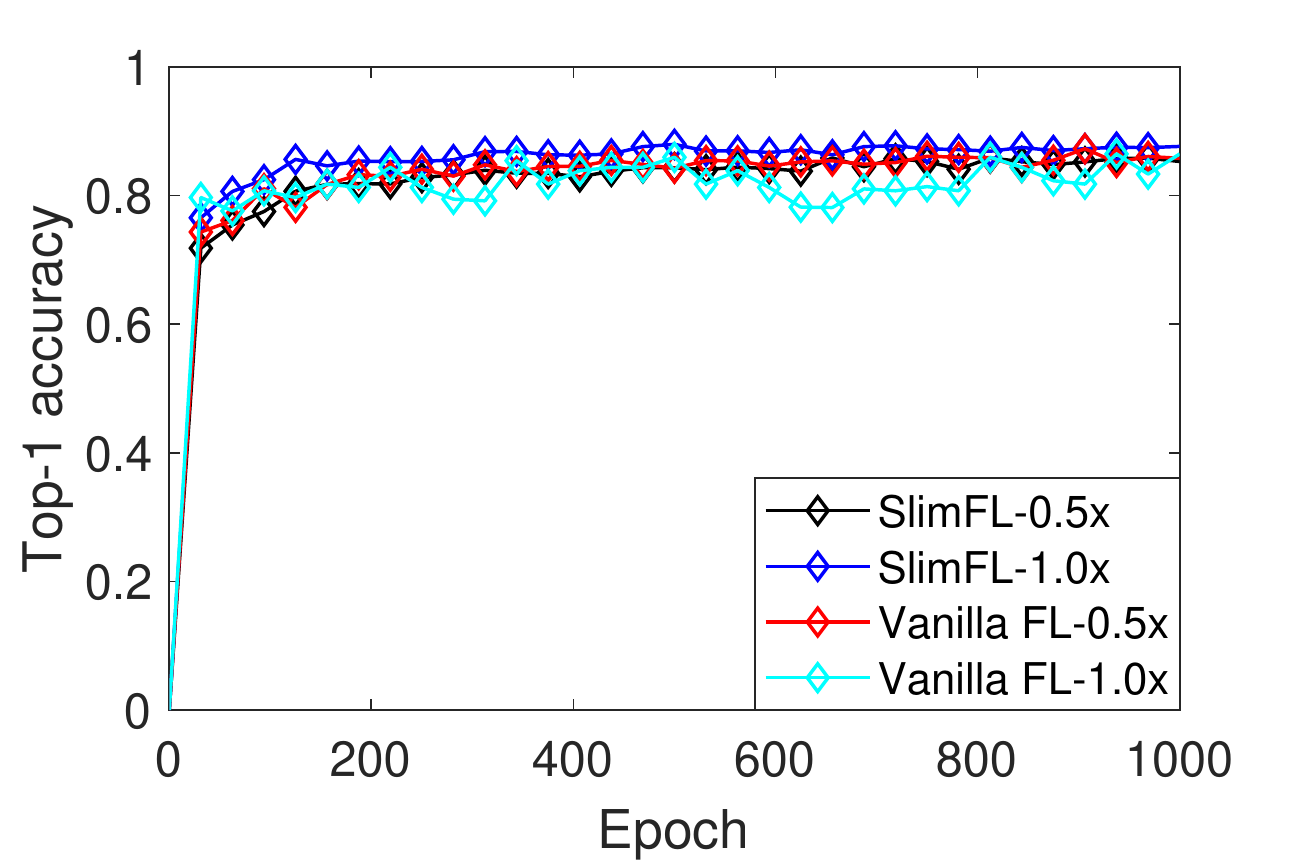}\\
    (a) $\alpha = 0.1$ & (b) $\alpha = 1.0$ & (c) $\alpha = 10$ \\ 
\end{tabular}
    \vspace{-2.5mm}
    \caption{{Test} accuracy in poor channel conditions (on average).}
    \label{fig:performance-moderate}
\end{figure*}

\BfPara{Robustness to Poor Channels}
Fig.~\ref{fig:performance-good} and Tab.~\ref{tab:accuracy} show that both SlimFL and Vanilla FL achieve high accuracy in good channel conditions. However as the channel condition deteriorates from good to poor channels, Fig.~\ref{fig:performance-moderate} and Tab.~\ref{tab:accuracy} illustrate that the maximum accuracy of Vanilla FL-1.0x at $\alpha=10$ drops from $86\%$ to $82\%$. Meanwhile, the accuracy of Slim-FL-1.0x keeps the same maximum accuracy $87$\% at $\alpha=10$ under both good and poor channels. What is more, at $\alpha=0.1$, SlimFL-1.0x even achieves $18$\% higher top-1 accuracy than Vanilla FL-1.0x that consumes more communication and computing costs. Furthermore, the std of Vanilla FL-1.0x's top-1 accuracy increases by up to $59$\% as channel condition deteriorates, whereas that of SlimFL increases by only up to $31$\%. These results advocate the robustness of SlimFL against poor channels, as well as its robustness to non-IID data distributions (i.e., low $\alpha$) and communication efficiency, as elaborated next.

\BfPara{Robustness to Non-IID data}
As illustrated in Fig.~\ref{fig:performance-moderate} and Tab.~\ref{tab:accuracy}, SlimFL-0.5x shows a stable convergence under the conditions of $\alpha$. Vanilla FL-0.5x and Vanilla FL-1.0x exhibit the std of 8.3 and 9.2, in poor communication condition and with the non-IID dataset ($\alpha =0.1$). On the contrary, both SlimFL-0.5x and SlimFL-1.0x exhibit the std of 2.4 and 2.9 at top-1 accuracy respectively. This tendency holds even when $\alpha = 1$, $\alpha = 10$. SlimFL-1.0x and SlimFL-0.5x exhibit lower variation than Vanilla FL-1.0x and Vanilla FL-0.5x. This underscores the robustness of SlimFL to non-IID data distributions in poor channels.

\BfPara{Communication Efficiency} 
The total amount of transmitted bits between 10 devices and server in ideal channel conditions (i.e., always successful decoding) are 205.8MBytes for SlimFL and Vanilla FL-1.0x, and 102.9MBytes for Vanilla FL-0.5x.
Tab.~\ref{tab:bits} shows that SlimFL achieves up to $3.52$\% less dropped bits than Vanilla FL-1.0x, thanks to the use of SC and SD. The reduced dropped bits of SlimFL can be found by the successfully decoded bits of 0.5x models that cannot be simultaneously received under Vanilla FL-1.0x. Note that SlimFL decodes less 1.0x model bits than Vanilla FL-1.0x, as a part of transmission power of SlimFL is allocated to 0.5x models. In return, SlimFL not only receives 1.0x models but also 0.5x models simultaneously. The additionally received 0.5x models correspond to the LH parts of the 1.0x models, which therefore improve the accuracy and convergence speed of both 0.5x and 1.0x models. SlimFL enjoys the aforementioned benefits while consuming only the half of the transmission power and bandwidth compared to Vanilla FL-1.5x, as illustrated in Tab.~\ref{tab:bits}, corroborating its communication efficiency.

\begin{table}[t!]
    \small\centering
    \resizebox{\linewidth}{!}{\begin{tabular}{c||ccc|ccc}
        \toprule[1pt]
        \multirow{3}{*}{\bf{Method}} &\multicolumn{6}{c}{\bf{Top-1 Accuracy (\%)}} \\ &\multicolumn{3}{c}{\bf{Good}}  & \multicolumn{3}{c}{\bf{Poor}}\\
         & $\alpha=0.1$ & $\alpha=1$ &  $\alpha=10$ & $\alpha=0.1$ & $\alpha=1$ &  $\alpha=10$ \\ \midrule
        SlimFL-0.5x & $54 \pm 2.2$ &  $ 83 \pm 1.0$ & $ 85 \pm 1.0$ & $56 \pm 2.4$ & $82 \pm 1.7$ & $ 85 \pm 1.1$ \\ 
        SlimFL-1.0x & $ 59  \pm  2.3$ &  $ 85 \pm 1.1$ & $ 87 \pm 1.1$ & $ 65 \pm 2.9$ & $ 84 \pm 1.4$ & $ 87 \pm 0.9$  \\
        Vanilla FL-0.5x & $ 45 \pm 5.9$ &  $ 84 \pm 1.1$ & $ 85 \pm 1.0$ & $ 39 \pm 8.3$ & $83 \pm 1.2$ & $ 85 \pm 0.9$\\
        Vanilla FL-1.0x & $69  \pm 5.8$ &  $ 85 \pm 4.0$ & $86  \pm  4.3$ & $ 55 \pm 9.2$ & $80 \pm  6.0$ & $ 82 \pm 4.7$ \\ \bottomrule[1pt]
    \end{tabular}}
    \caption{Accuracy under different channel conditions and~$\alpha$.}\vspace{-3mm}
    \label{tab:accuracy}
\end{table}

\begin{table}[t!]
    \small\centering
    \resizebox{\columnwidth}{!}{\begin{tabular}{c|ccc|cc|cc}
        \toprule[1pt]
        \bf Decoding Success & \multicolumn{3}{c|}{\bf SlimFL} & \multicolumn{2}{c|}{\bf Vanilla FL-0.5x}  & \multicolumn{2}{c}{\bf Vanilla FL-1.0x}  \\
        \bf Bits [MBytes] &0.5x& 1.0x & drop & 0.5x & drop & 1.0x & drop \\ \midrule
        Uplink (Good)  & 1.96 & 198.45 & 5.46 & 102.21 & 0.72 & 200.30 & 5.56  \\ 
        Downlink (Good) & 0.52 & 204.01 & 1.34 & 102.72 & 0.21 &  204.42 & 1.44\\
        Uplink (Poor)   & 18.32 & 130.10 & 57.44 &93.87 & 9.06 & 144.93 & 60.96  \\ 
        Downlink (Poor) & 5.87 & 183.42 & 16.57 & 100.56 & 2.37 &   188.57 & 17.29\\ \bottomrule[1pt]
    \end{tabular}}
    \vspace{-2mm}
    \caption{Successfully decoded bits of SlimFL, Vanilla FL-0.5x, and Vanilla FL-1.0x.}
    \label{tab:bits}
    \vspace{-3mm}
\end{table}

\BfPara{Energy Efficiency}
Thus far we have measured the performance of SlimFL after training with a fixed $1000$ epochs. Here, we measure the energy expenditure until convergence, where the training convergence is defined by the moment when the standard deviation (std) of test accuracy is below a target threshold and the minimum test accuracy becomes higher than the average test accuracy in $100$ consecutive rounds (see the details in Appendix~\ref{sec:appendix_H}). Given the communication and computing energy costs per round in Tab.~\ref{tab:energy}, Tab.~\ref{tab:energy2} compares the total energy costs of SlimFL and Vanilla FL-1.5x until convergence. The results show that on average, SlimFL achieves $3.6$x less total computing cost with $2.9$x lower total communication cost until convergence. Such higher energy efficiency comes from the faster convergence of SlimFL even under non-IID distributions and/or poor channel conditions thanks to SC and SD.

\begin{table}[t!]
    \small\centering
    \resizebox{\columnwidth}{!}{\begin{tabular}{c|c|c}
        \toprule[1pt]
        {\bf{Metric}} & \bf{SlimFL} & \bf{Vanilla FL-1.5x} \\\midrule
         Communicational Cost [mW/Round]           & 125           & 250   \\
        {Computational Cost [MFLOPS/Epoch]} & {$3.56$} & $3.56$\\\bottomrule[1pt]
    \end{tabular}}
    \caption{Transmission energy and computational cost per communication round.}
    \label{tab:energy}
    \vspace{-5mm}
\end{table}

\begin{table}[t!]
    \small\centering
    \resizebox{\columnwidth}{!}{\begin{tabular}{c|c||cc|cc}\toprule[1pt]
        \multirow{2}{*}{\bf{Metric}} &  \multirow{2}{*}{\bf{non-IIDness}} & \multicolumn{2}{c|}{\bf{SlimFL}} & \multicolumn{2}{c}{{\bf Vanilla FL-1.5x}}\\
         &  &  \bf{Good} & \bf{Poor} &  \bf{Good} & \bf{Poor}\\\midrule
         
        \multirow{3}{*}{\shortstack{Communicational \\ Cost [W]}} & \bf{$\alpha=0.1$} & 44.5 & 35.9 & 99.5 & 123.3 \\
         & \bf{$\alpha=1.0$} & 5.3 & 6.5 & 9.9 & 23.0 \\
         & \bf{$\alpha=10$}  & 1.9 & 2.2 & 6.4 & 15.9 \\\midrule
         
        \multirow{3}{*}{\shortstack{Computational \\ Cost [GFLOPS]}} & \bf{$\alpha=0.1$} & 1.27 & 1.02 & 1.88 & 2.41 \\
         & \bf{$\alpha=1.0$} & 0.15 & 0.18 & 0.22 & 0.51\\
         & \bf{$\alpha=10$} & 0.05 & 0.06 & 0.14 & 0.35\\ \bottomrule[1pt]
     \end{tabular}}
    \caption{Total computational cost and transmission power of SlimFL and Vanilla FL-1.5x in good/poor channel condition with various non-IIDness ($\alpha=0.1,1.0,10)$.}
    \label{tab:energy2}
    \vspace{-5mm}
\end{table}\section{Conclusion}\label{sec:6}
FL is a promising solution to enable high-quality on-device learning while protecting user privacy. However, existing FL solutions cannot cope flexibly with different devices having heterogeneous levels of available energy and channel throughput, without significantly compromising communication and energy efficiencies.

To tackle this problem, we propose a novel FL framework that integrates SC and SD communication methods with a width-adjustable SNN architecture. Extensive experiments verify that SlimFL is a communication and energy efficient solution under various communication environments and data distributions. Particularly under poor channel conditions and non-IID data distributions, SlimFL even achieves higher accuracy and faster convergence as well as lower energy expenditure than its vanilla FL counterpart consuming $2$x more communication resources. Studying the impact of more adjustable SNN width levels could be an interesting topic for future work. Another interesting direction is to apply SlimFL for multitask learning in which various width configurations correspond to different tasks.

\section*{Acknowledgment}
This research was funded by Ministry of Health and Welfare (HI19C0842) and also by Academy of Finland6G Flagship (grant no.318927), project SMARTER, projects EU-ICT IntellIoT and EUCHISTERA LearningEdge, Infotech-NOOR and NEGEIN.

\bibliographystyle{icml2021}

\newpage
\vfill

\pagebreak
\appendix

\section*{Appendices}

\section{Local SNN Training Algorithm}\label{sec:Appendix-1}

\begin{figure}[t!]
    \centering
    \includegraphics[width=\columnwidth]{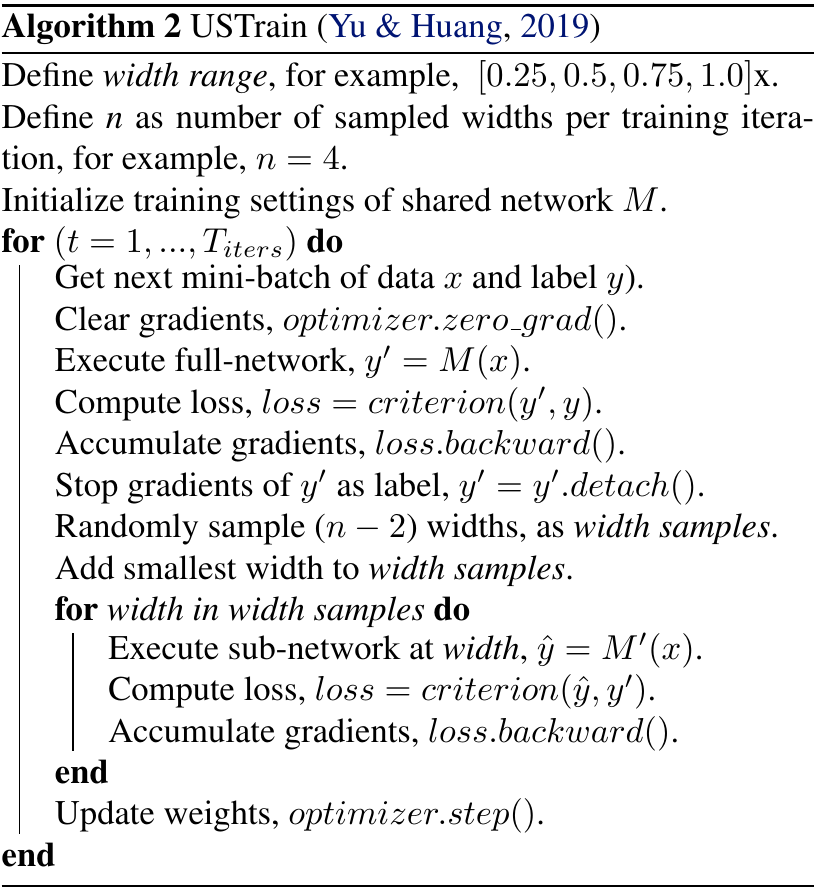}
\end{figure}
  
\begin{figure}[ht]
\includegraphics[width=1\linewidth]{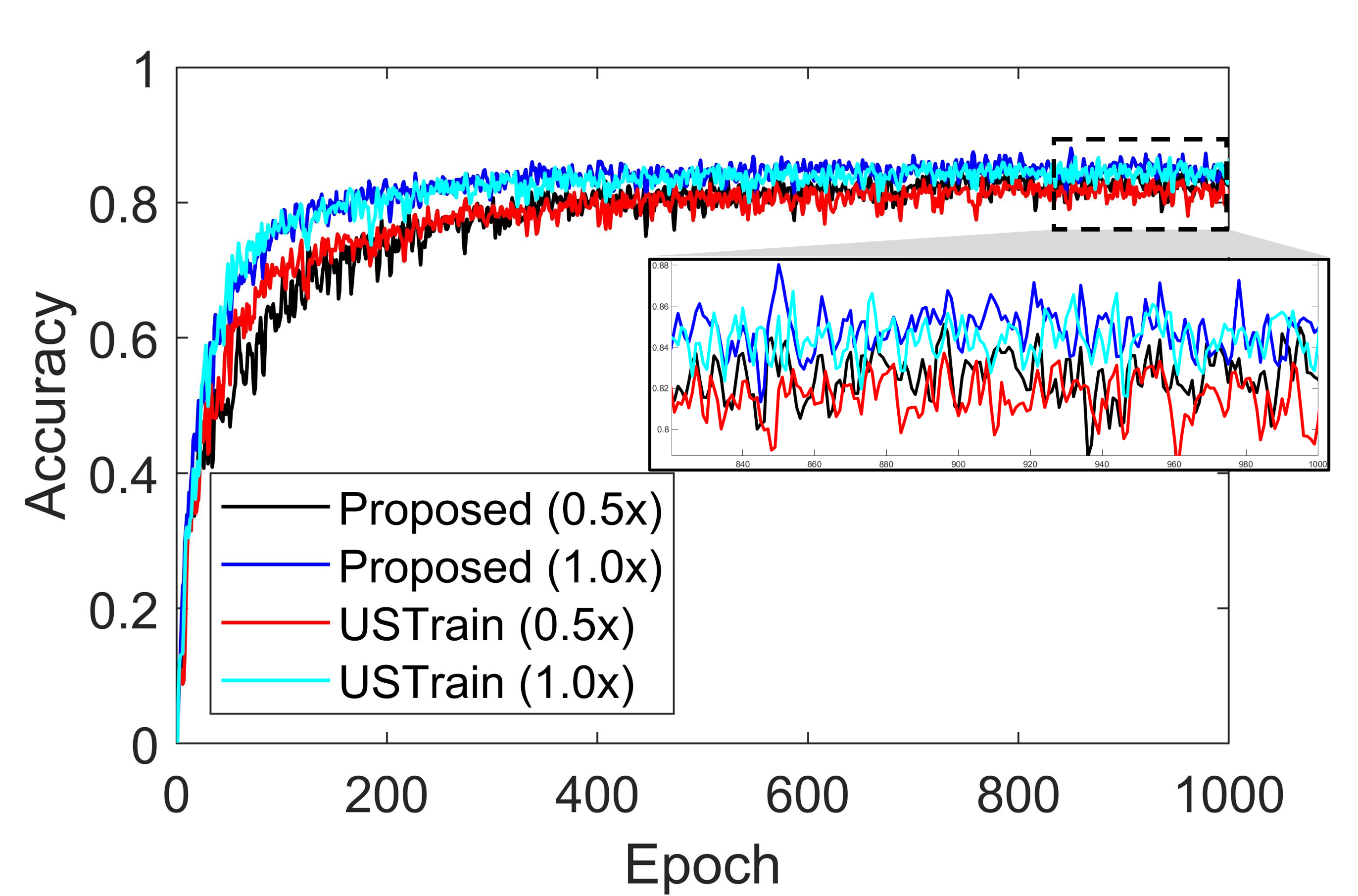}
\caption{Comparison of two different training methods in poor channel conditions with $\alpha=1$}
\label{fig:train}
\end{figure}

The state-of-the-art SNN training algorithm is the USTrain method in Algorithm~2 proposed by \cite{ICCV2019_USlimmable}. At its core, USTrain utilizes the inplace distillation and the sandwich rule that are effective under the case where an SNN can be divided into more than two segments. The SNN architecture under study consists of only the LH and RH segments, making the Algorithm~2 unfit for our case. Alternatively, we propose a new weighted-sum based SNN training algorithm in \eqref{eq:localtrain}, which is inspired by the training algorithm used in depth-controllable model training \cite{AAAI2019_Anytime}.
Fig.~\ref{fig:train} shows our proposed local training method outperforms USTrain when training an UL-MobileNet SNN model.

\section{Optimizer}
\begin{figure}[ht]
\includegraphics[width=1\linewidth]{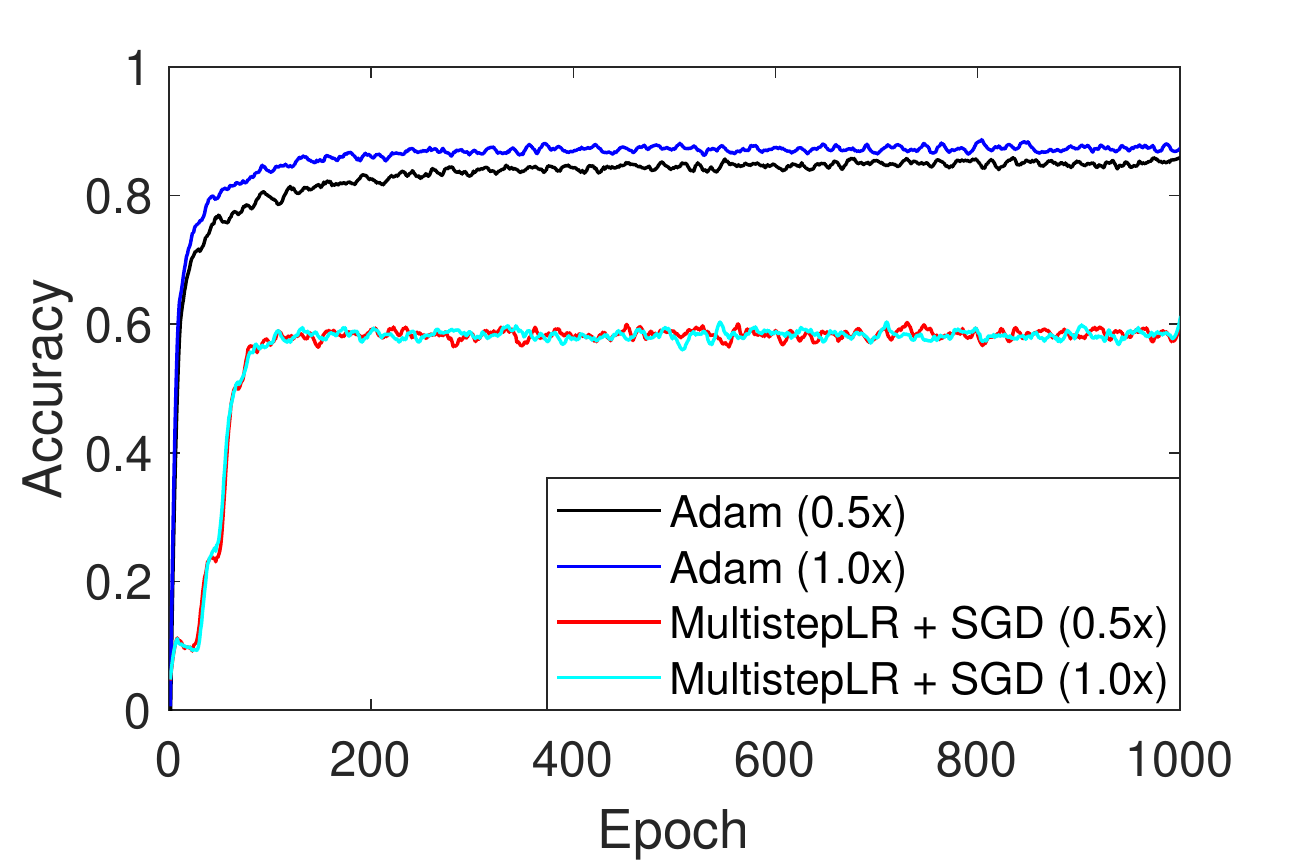}
\caption{Multi-step-LR + SGD Optimizer in poor channel condition with $\alpha=1$}
\label{fig:optim}
\end{figure}
The optimizer of universally slimmable neural network consists of stochastic gradient descent optimizer with multi-step learning rate scheduler~\cite{ICCV2019_USlimmable}. However, UL-MobileNet converges to local minima when using multi-step-LR + SGD optimizer as shown in Fig.~\ref{fig:optim}. Therefore, we adopt Adam optimizer as an optimizer of UL-MobileNet.

\section{Dataset} \label{sec:appendix_dataset}
\begin{figure}[ht]
\includegraphics[width=1\linewidth]{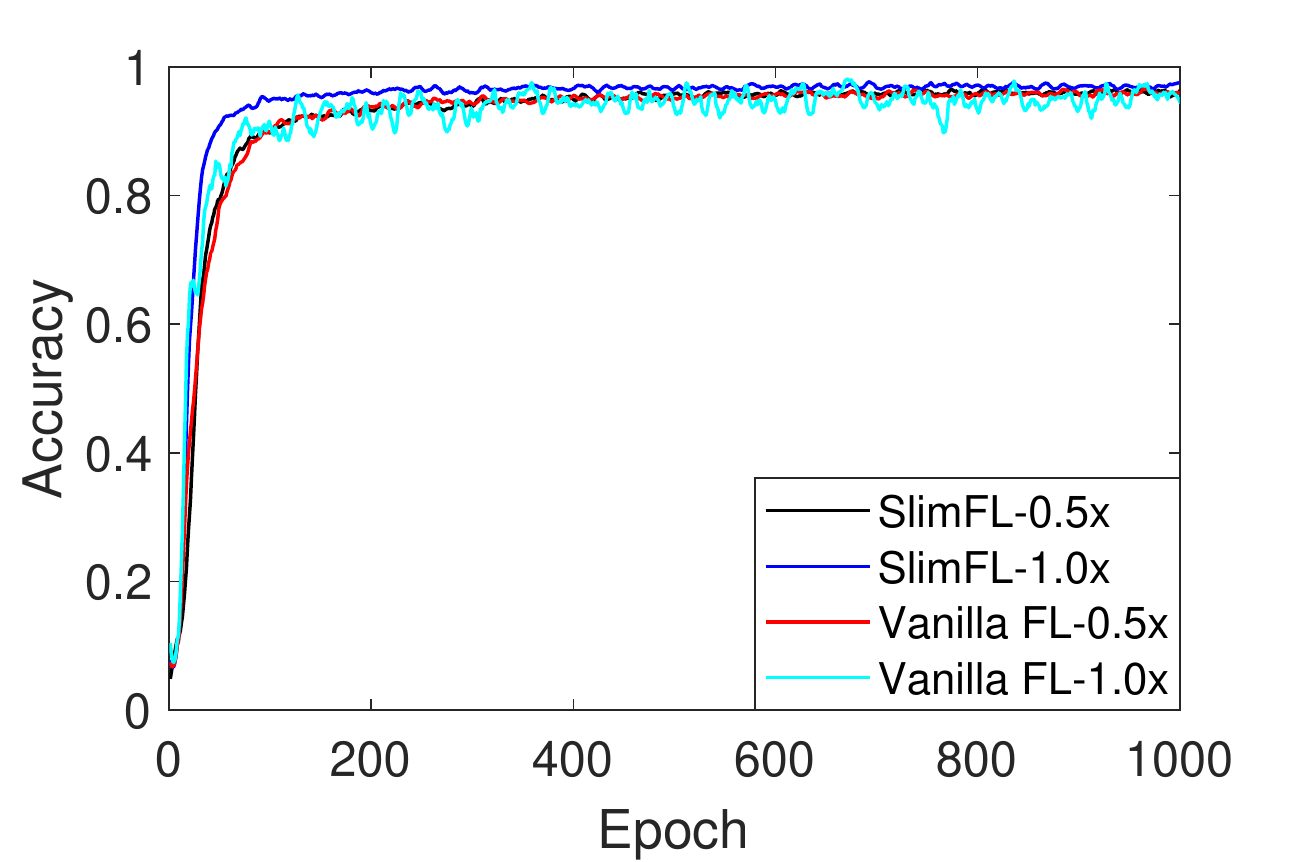}
\caption{Performance of proposed algorithm and two comparisons with MNIST dataset ($\alpha=1$) in poor channel condition.}
\label{fig:MNISTperformance}
\end{figure}

In this paper, as Fig.~\ref{fig:MNISTperformance} shows, all models exhibit high accuracy and reasonable convergence when using MNIST dataset. Therefore we use Fashion MNIST dataset to show the differences between models in the main experiment.
The result of accuracy is too high as shown in Fig.~\ref{fig:MNISTperformance}. With MNIST dataset, the performance difference between 0.5x and 1.0 is too trivial so we adopt Fashion MNIST dataset for UL-MobileNet.
\begin{figure*}[t!]
\small\centering
\begin{tabular}{ccc}
    \includegraphics[width=0.31\linewidth]{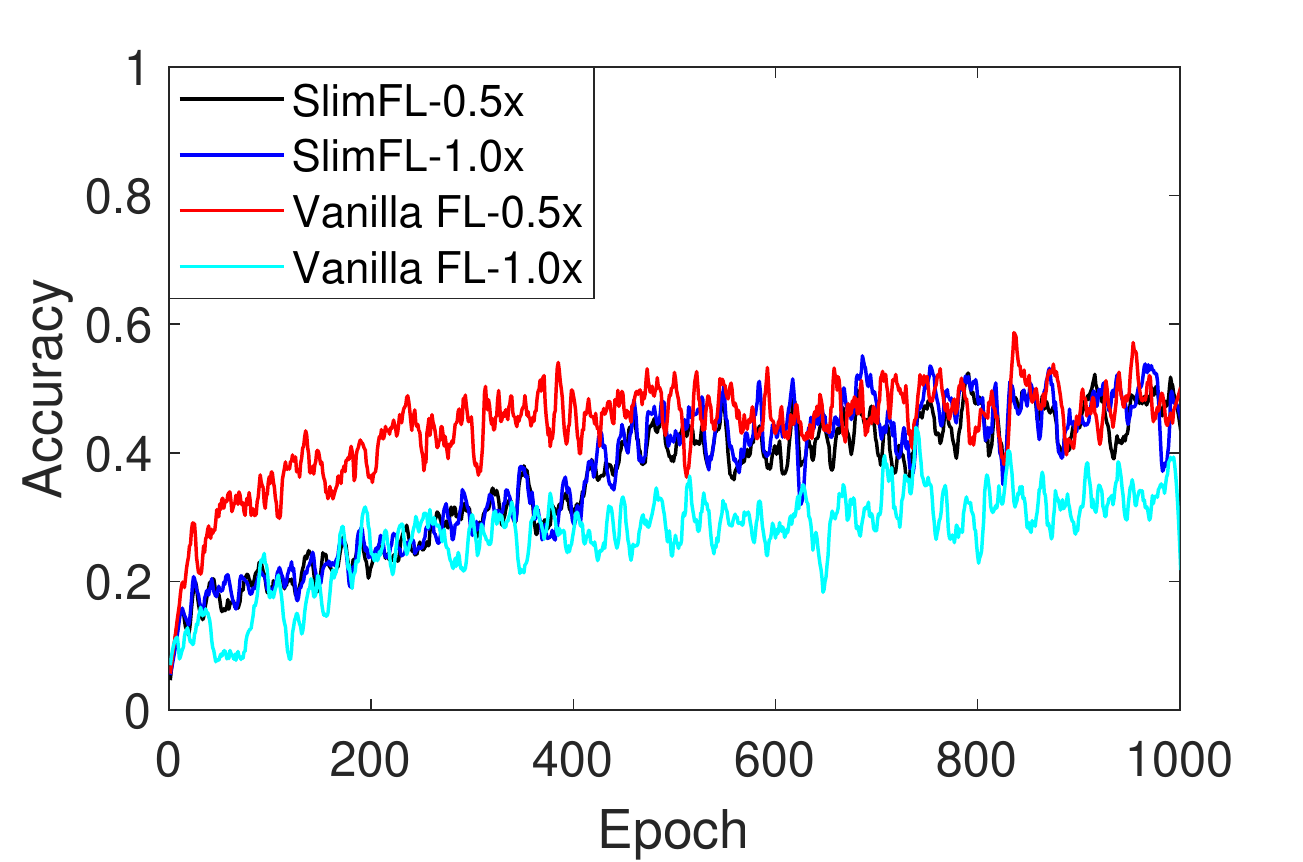} & \includegraphics[width=0.31\linewidth]{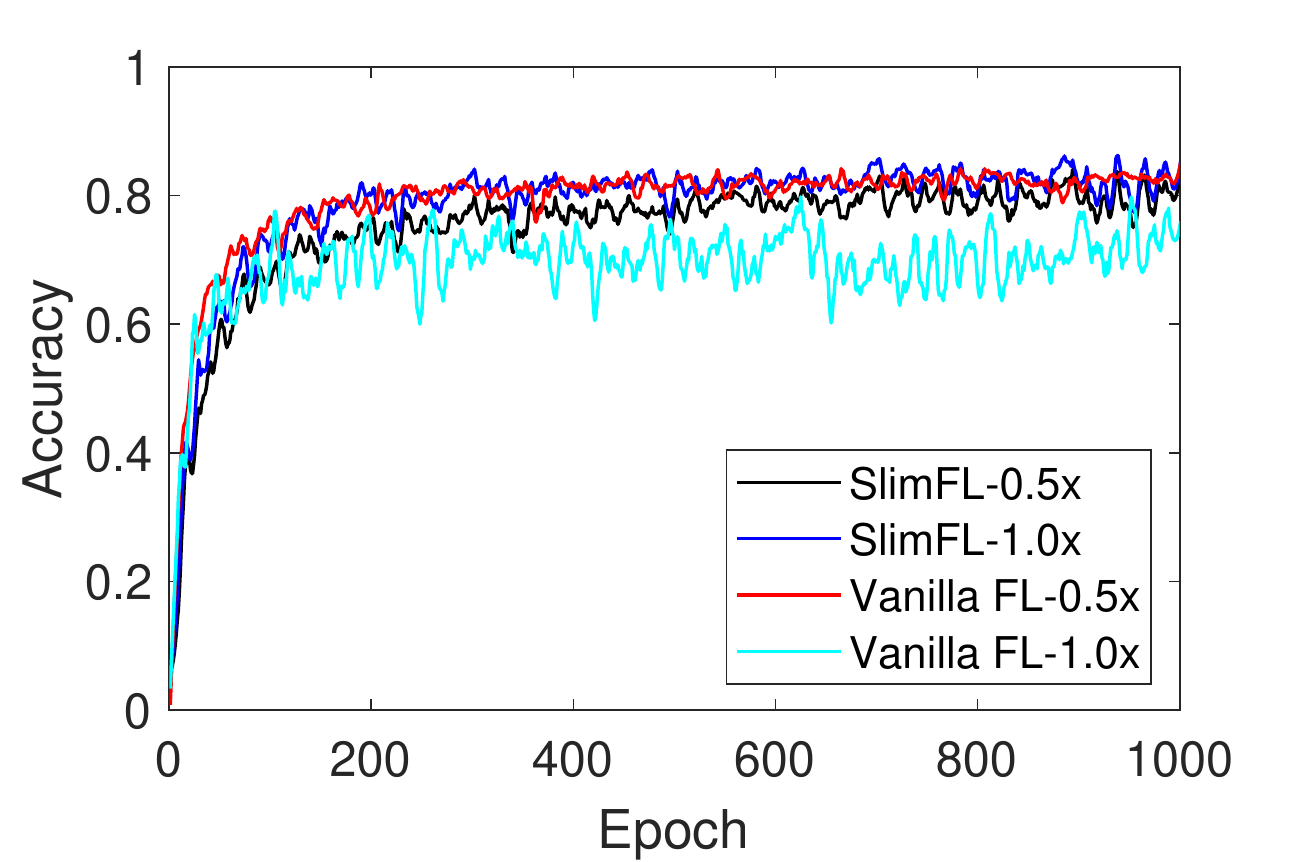} & 
    \includegraphics[width=0.31\linewidth]{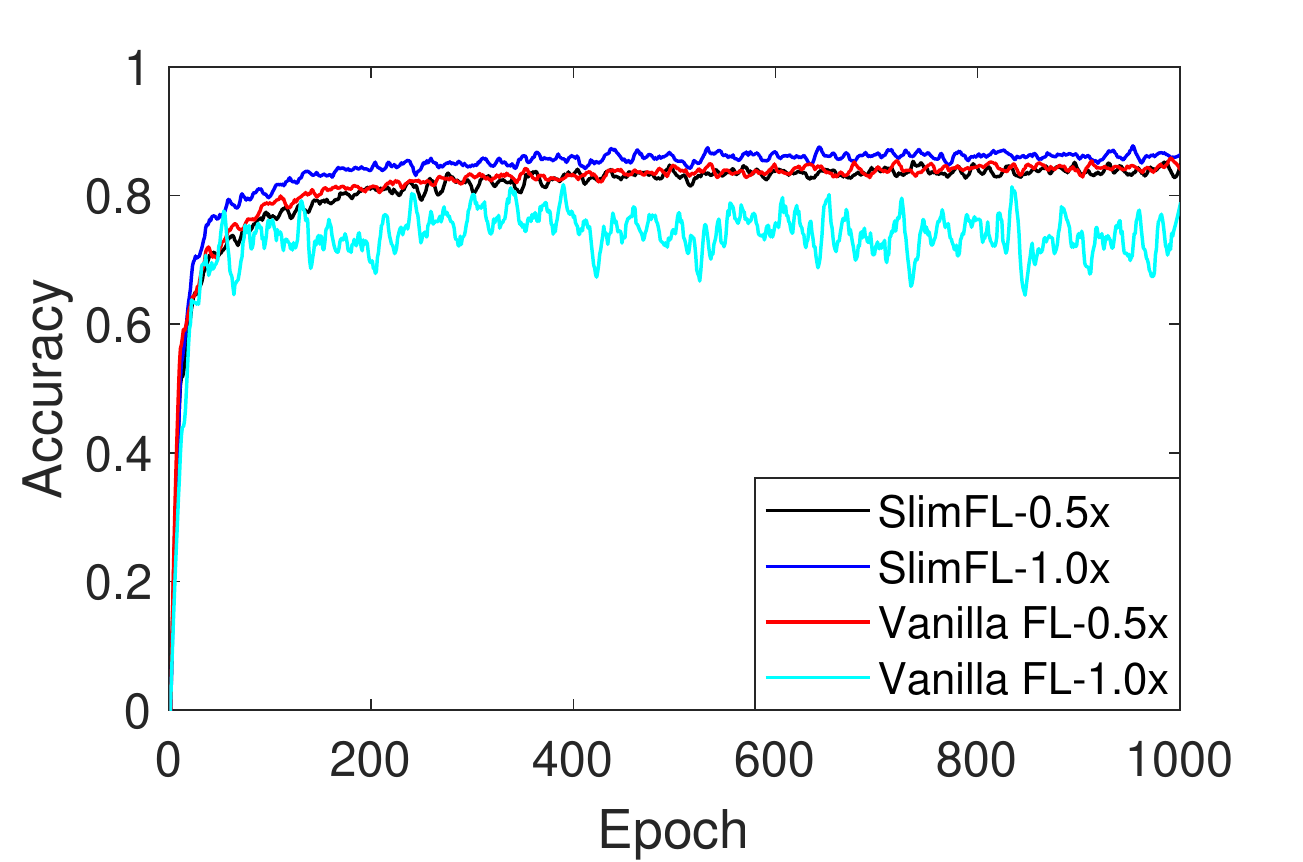}\\
    (a) $\alpha = 0.1$ & (b) $\alpha = 1.0$ & (c) $\alpha = 10$ \\ 
\end{tabular}
    \caption{Performance difference in extremely poor channel conditions.}
    \label{fig:performance-poor}
\end{figure*}

\section{Decoding Success Probability Derivation} \label{sec:appendix_dsp}
The distribution of the throughput $R_k$ for the $k$-th message is cast as
\begin{align}
\Pr(R_k \geq t)
&= \Pr\left( W \log_2\left(1 + \frac{g d^{-\beta}P_k^T}{\sigma^2 + P_k^I} \right) \geq t \right)
\\
&= \Pr(\frac{g d^{-\beta} P^T_k }{\sigma^2 + P^I_k} \geq t'), \label{eq:dsp}
\end{align}
where $t'=2^{t/W}-1$. Here, $P^I_k = g d^{-\beta} \hat{P}_k^I$, with $\hat{P}_k^I:=\sum^{K}_{k' = k+1}P^T_{k'}$ for $k\leq K-1$, and $\hat{P}^I_K=P^I_K=0$. By applying this, \eqref{eq:dsp} is recast as
\begin{align}
\eqref{eq:dsp} 
&=  \Pr \left( g d^{-\beta}P_k^T \geq t'\left(  \sigma^2 + g d^{-\beta}\hat{P}^I_k \right)   \right) \\
&= \Pr\left(g \geq \frac{c}{ P^T_k/t' - \hat{P}^I_k}      \right),
\end{align}
where $c=\sigma^2 d^\beta$.
Applying this result, the decoding success probability $p_k$ of the $k$-th message is represented as
\begin{align}
    &p_k 
    = \Pr(R_1 \geq t, R_{2} \geq t,\cdots, R_k \geq t  )\\
    &= \Pr \!\left( g\geq \frac{c}{ P^T_1/t' - \hat{P}^I_1}, \cdots , g \geq \frac{c}{ P^T_k/t' - \hat{P}^I_k} \right) \\
    &= \Pr\!\left( g \!\geq\!  \max \!\left\{\!
    \frac{c}{ P^T_1/t' - \hat{P}^I_1} , \cdots\!, \frac{c}{ P^T_k/t' - \hat{P}^I_k} 
    \!\right\}\!\right)\!. \label{eq:sicdsp1}
\end{align}

For $K=2$, $\hat{P}^I_1=P_2^T$ and $\hat{P}^I_2=0$. Following \eqref{eq:sicdsp1} for $P_1^T \gg P_2^T$ with SD, the decoding success probabilities of the two messages are given as follows:
\begin{align}
p_1 &= \Pr\left( g \geq \frac{c}{P_1^T/t' - P_2^T}  \right) \label{eq:p1}    \\
P_2&= \Pr\left( g \geq \frac{c}{P_2^T/t'}  \right) \label{eq:p2}
\end{align} 

Under Rayleigh fading, the small-scale fading power gain $g$ follows an exponential distribution, i.e., $g\sim\textsf{Exp}(1)$. Applying the complementary cumulative distribution function (CCDF) of an exponential random variable to \eqref{eq:p1} and \eqref{eq:p2}, the decoding success probabilities finally become:
\begin{align}
p_1 &= \exp\left( - \frac{c}{P_1^T/t' - P_2^T} \right)  \label{eq:p1exp}\\ 
p_2 &= \exp\left(- \frac{c}{P_2^T/t'}  \right). \label{eq:p2exp}
\end{align}

\section{Convergence Measurement} \label{sec:appendix_H}
We design a method to measure the convergence of each model in all the experimental environments presented in Tab.~\ref{tab:energy2}. Tab.~\ref{tab:accuracy} represents that SlimFL and Vanilla FLs (i.e., 0.5x and 1.0x) have huge differences in std and mean of Top-1 accuracy in identical environments. These differences between SlimFL and Vanilla FLs make it difficult to set common criteria for measuring convergence as shown in Tab.~\ref{tab:accuracy}. Therefore, to measure the convergence of models, we define the reference values of the mean ($mean_{ref}$) and std ($std_{ref}$), respectively. To describe in details, we set $mean_{ref}$ as 80\% of the last 100 epoch's average Top-1 accuracy and $std_{ref}$ as 7.2\%. We define the convergence when average of Top-1 accuracy for 100 consecutive epochs is higher than $mean_{ref}$, and the average std is lower than $std_{ref}$.  

\section{Extremely Poor Communication Channels}\label{sec:Appendix-2}
\begin{table}[th]
    \small\centering
    \resizebox{\columnwidth}{!}{\begin{tabular}{c|cccccccc}
        \toprule[1pt]
        \multirow{2}{*}{Condition} & \multicolumn{8}{c}{Decoding Success Probability}   \\
         & $p^{up}_1$ & $p^{up}_2$ &  $p^{up}_{cp1}$ & $p^{up}_{cp2}$ & $p^{dn}_{1}$ & $p^{dn}_{2}$& $p^{dn}_{cp1}$ & $p^{dn}_{cp2}$ \\ \midrule
        Extremely Poor & 0.22 & 0.08 & 0.50 & 0.07 & 0.80 & 0.64 & 0.86 & 0.56 \\ 
        Uplink SC+SD/Downlink 1.0x  & 0.76 & 0.70 & 0.92 & 0.38 & - & 0.74 & 0.97 & 0.74 \\ 
        Uplink SC+SD/Downlink 0.5x & 0.76 & 0.70 & 0.92 & 0.38 & 0.97 & - & 0.97 & 0.74 \\ 
        Uplink 0.5x/Downlink SC+SD & - & 0.38 & 0.92 & 0.38 & 0.97 & 0.94 & 0.97 & 0.74 \\ 
        \bottomrule[1pt]
    \end{tabular}}
    \caption{Decoding success probabilities under an extremely poor channel condition and transmission method.}
    \label{tab:probability-poor}
\end{table}

We simulate the performance of our proposed model in extremely poor channel conditions as shown in Tab.~\ref{tab:probability-poor}. Fig.~\ref{fig:performance-poor} presents the learning curve in respect to accuracy of three algorithms (i.e., Proposed, Comp1, and Comp2). 
With non-IID dataset ($\alpha=0.1$), Fig.~\ref{fig:performance-poor}(a) shows that the accuracy of {SlimFL} (both 0.5x and 1x) converges to 0.44 at 620 epochs. Also Comp1 and Comp2 converge to 0.44 and 0.38 at 380 epochs and 910 epochs, respectively. As shown in Fig.~\ref{fig:performance-poor}(b), the accuracy of {SlimFL} and Comp1 converge to 0.81 and Comp2 converges to 0.76. Finally, using dataset close to IID (i.e., $\alpha=10$), {SlimFL} 1.0x shows the highest accuracy, {SlimFL} 0.5x and Comp1 follow next, and Comp2 shows the lowest accuracy. SlimFL outperforms Comp1 and Comp2 when $\alpha=1$ and $\alpha=0.5$ regarding accuracy even in extremely poor channel condition.

\section{SC and SD in the Uplink and/or Downlink} \label{sec:appendix_E}
\begin{figure}[ht]
\includegraphics[width=1\linewidth]{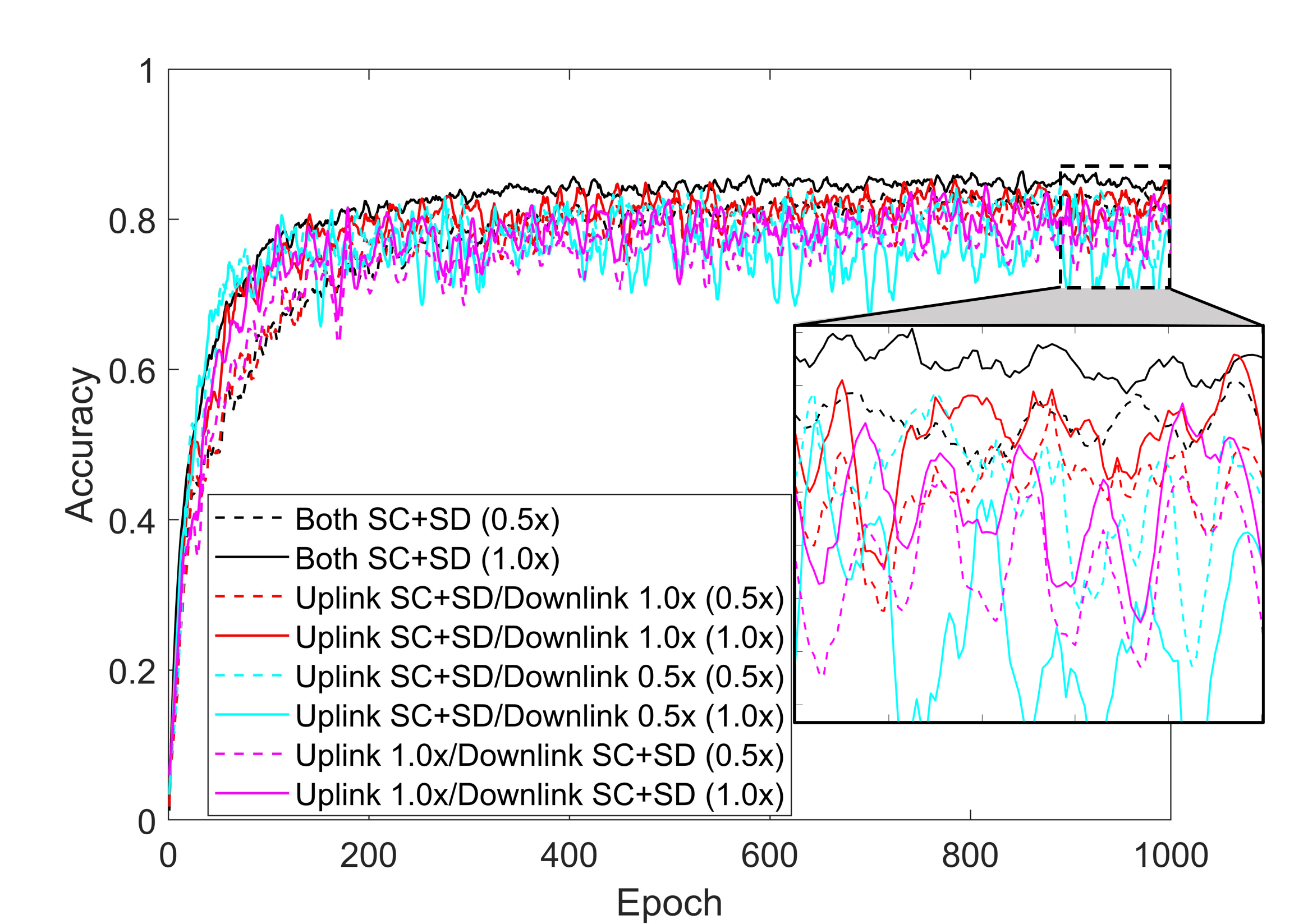}
\caption{Performance difference in various transmission method (Poor, $\alpha=1$).}\label{fig:noma}
\end{figure}
We design our proposed algorithm with utilizing SC and SD in both uplink and downlink communication. We simulated with various transmission methods (e.g. Uplink SC+SD/Downlink 1.0x, Uplink SC+SD/Downlink 0.5x, Uplink 1.0x/Downlink SC+SD). The first comparison (Uplink SC+SD/Downlink 1.0x) is that devices upload both $\theta_{1/2}$ and $\theta_{2/2}$ with SC and SD, and the server transmits 1.0x to devices. The second comparison  (Uplink SC+SD/Downlink 0.5x) is that devices upload both $\theta_{1/2}$ and $\theta_{2/2}$ and the server transmits $\theta_{1/2}$ to devices. The third comparison (Uplink 1.0x/Downlink SC+SD) is that devices can transmit only $\theta$ and the server can transmit both $\theta_{1/2}$ and $\theta_{2/2}$ to local model. Since Uplink 0.5x/Downlink SC+SD is not proper transmission technique, we do not consider Uplink 0.5x/Downlink SC+SD. The decoding success probability of three techniques are shown in Tab.~\ref{tab:probability-poor}. 

We simulated with those three comparison technique, and Fig.~\ref{fig:noma} shows the result. In Fig.~\ref{fig:noma}, learning curves converge with a similar tendency. At the end of training phase (i.e., 800 -- 1000 epochs), the accuracy is found to be high in the order of our proposed method (Both SC+SD), Uplink SC+SD/Downlink 1.0x, Uplink 1.0x/Downlink SC+SD, and Uplink 0.5x/Downlink SC+SD. Consequently, our proposed  wireless communication system (i.e., both 
SC+SD) shows the best performance among those comparisons.

\section{Scability} \label{sec:appendix_F}
\begin{figure}[t!]
\small\centering
    \includegraphics[width=1\linewidth]{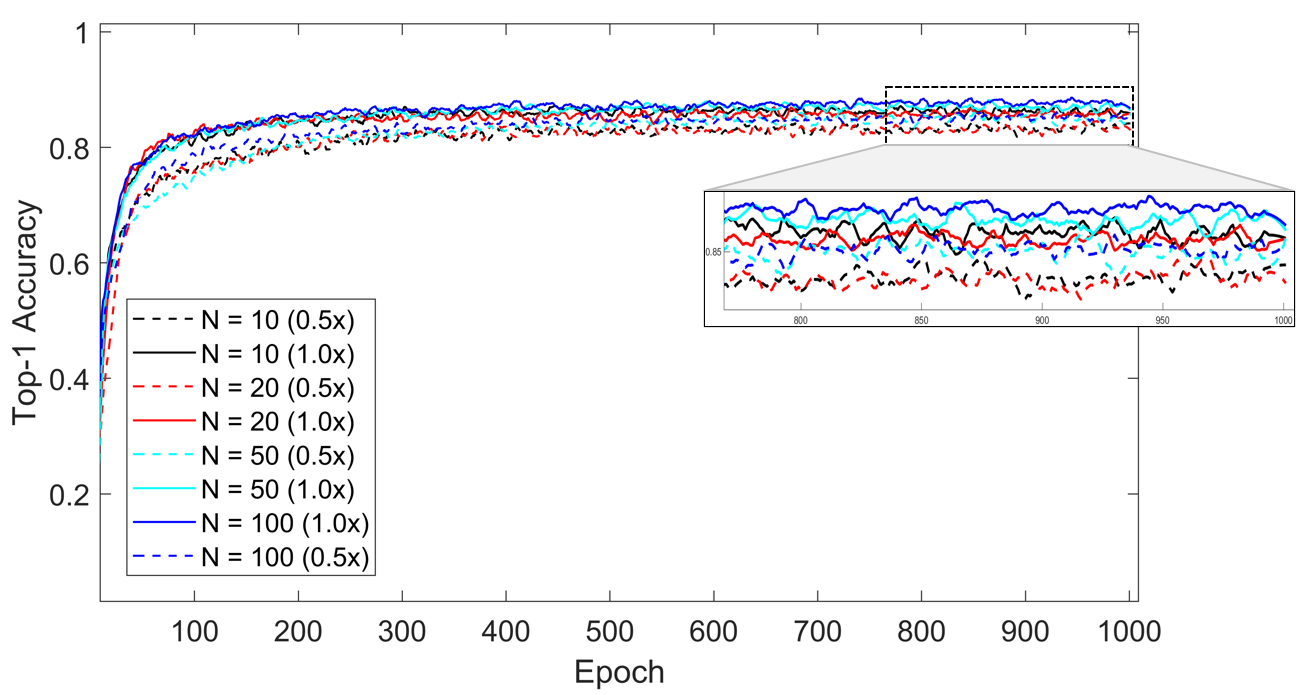}
    \caption{Top-1 accuracy with different number of devices.}
    \label{fig:Number of worker}
\end{figure}

Fig.~\ref{fig:Number of worker} illustrates the top-1 accuracy of SlimFL for a different number $N$ of devices. As $N$ grows from $10$ to $100$, the accuracy of the $0.5$x model increases from $84$\% to $86$\% that coincides with the accuracy of the $1.0$x model when $N=10$. In other words, federating with $10$x more devices is equivalent to running a $2$x larger model. For the $1.0$x model, the gain from federating more devices than $N=10$ is negligible. Studying such a scalability trened under different datasets and/or model architectures would be an interesting topic for future work.

\end{document}